\documentclass[preprint,12pt]{elsarticle}

\usepackage{amssymb}
\usepackage{amsmath}
\usepackage{amsfonts}
\usepackage{graphicx} 			
\usepackage{float} 				 
\usepackage{subcaption} 
\usepackage{subfig}		
\usepackage{threeparttable}
\usepackage{ragged2e}
\usepackage{bm}
\usepackage{algorithm, algpseudocode}
\usepackage{multirow}
\usepackage{url}
\usepackage{booktabs}

\begin{document}

\begin{frontmatter}



\title{M2I2HA: A Multi-modal Object Detection Method Based on Intra- and Inter-Modal Hypergraph Attention\tnoteref{t1}}
\tnotetext[t1]{This work is primarily supported by the National Science and Technology Major Project of China-Intelligent Manufacturing Systems and Robots (Grant No.2025ZD1603100 and 2025ZD1603101). It is also supported in part by the National Key R\&D Program of China (Grant No.2024YFE03250300).}

\author{Xiaofan Yang}
\ead{yxf97niu@stu.hit.edu.cn}

\author{Yubin Liu\corref{cor1}}
\ead{liuyubin@hit.edu.cn}

\author{Wei Pan}

\author{Guoqing Chu}

\author{Junming Zhang\corref{cor2}}
\ead{Junmingzhang@hit.edu.cn}

\author{Jie Zhao}

\author{Zhuoqi Man}

\author{Xuanming Cao}

\cortext[cor1]{Corresponding author}
\cortext[cor2]{Corresponding author}

\affiliation{organization={Harbin Institute of Technology}, %Department and Organization
            addressline={Xidazhi Street 92, Nangang District}, 
            city={Harbin},
            postcode={150001}, 
            state={Heilongjiang},
            country={China}}

%% Abstract
\begin{abstract}
Recent advances in multimodal detection have significantly improved detection accuracy in challenging environments. By integrating RGB with modalities such as thermal and depth, multimodal fusion increases data redundancy and system robustness. However, effectively extracting task-relevant features, achieving cross-modal alignment, and satisfying real-time constraints remain formidable challenges. Traditional Convolutional Neural Networks (CNNs) are limited by constrained receptive fields in capturing long-range dependencies. Transformer-based architecture suffers from quadratic computational complexity that hinders real-time deployment. State Space Models (SSMs) like Mamba offer linear efficiency but disrupt topological structures. To address these issues, this paper proposes a real-time multimodal detection network based on hypergraph theory, termed M2I2HA. An Intra-Hypergraph Enhancement module is developed to model high-order dependencies within each modality while improving parameter efficiency through low-rank decomposition and sparse hyperedge assignment. Meanwhile, an Inter-Hypergraph Fusion module exploits cross-modal hyperedge interactions to integrate complementary multimodal features and reduce the adverse effects of modality heterogeneity and spatial misalignment. A Multimodal and Multi-level Feature Dynamic Fusion Pipeline (M3FDFP) block is introduced to enable adaptive multi-level feature redistribution and alleviate data bottlenecks. Extensive experiments against state-of-the-art methods demonstrate that M2I2HA achieves competitive detection accuracy and fast inference speed. The code and pre-trained models will be available at: \url{https://github.com/WSYANGSX/machine_learning}.
\end{abstract}

\begin{graphicalabstract}
\includegraphics[width=\linewidth]{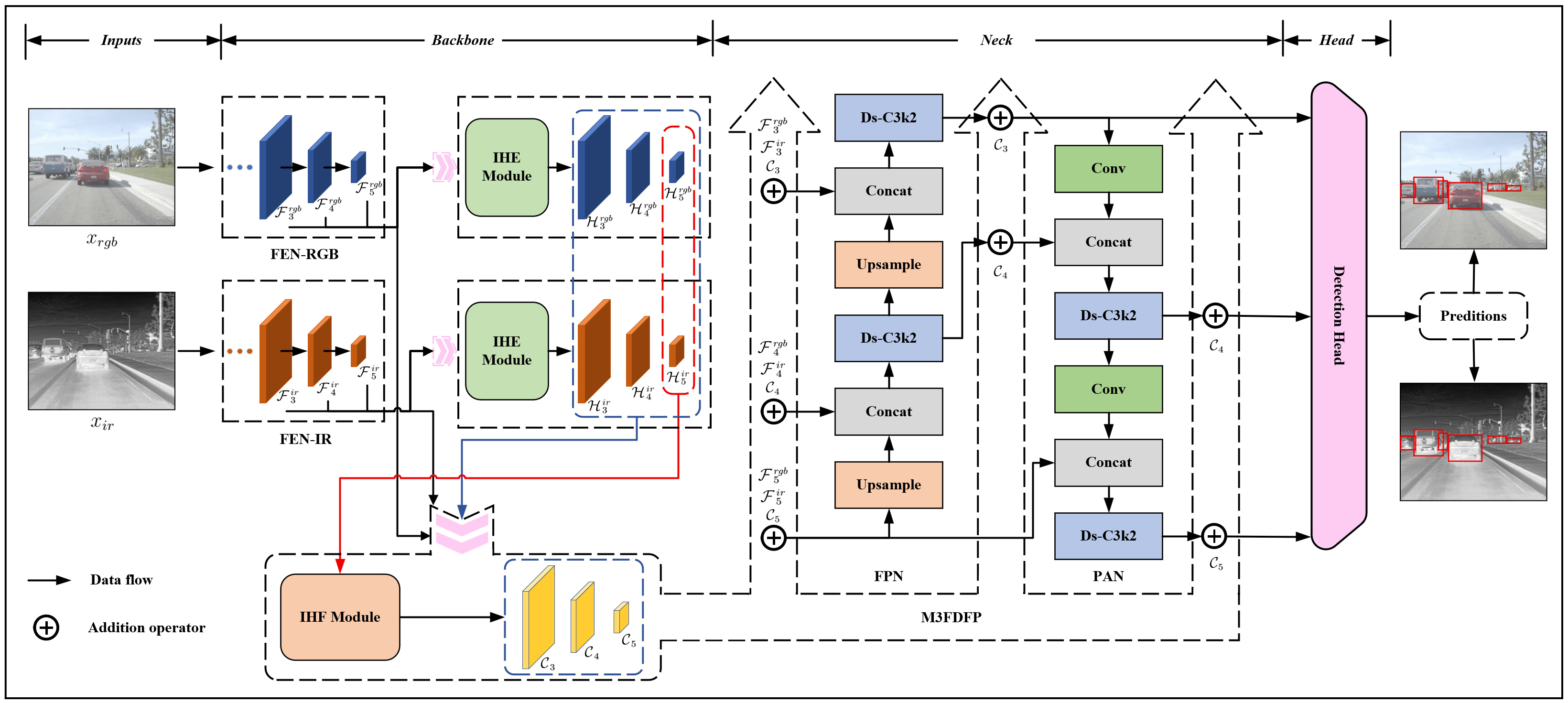}
\end{graphicalabstract}

\begin{highlights}
\item A novel multi-modal fusion object detection network named M2I2HA
\item Enhance detection accuracy and robustness in adverse visual conditions
\item An Inter-Hypergraph Fusion module to align and enhance cross-modal features
\end{highlights}

%% Keywords
\begin{keyword}
multimodal fusion \sep multimodal object detection \sep cross-modal attention \sep hypergraph
\end{keyword}

\end{frontmatter}



\section{Introduction}

Object detection, a critical downstream task in computer vision, has been widely applied in perception fields such as safe human-robot collaboration (HRC) \cite{dudek2024efficiency, robinson2023robotic}, vision-guided sorting and assembly \cite{konstantinidis2023multi}, autonomous navigation for AGVs \cite{patruno2024vision, xu2023onboard}, and drone-based industrial asset monitoring \cite{zhai2023yolo, lim2023heat}. Researchers have successively proposed a variety of detection algorithms \cite{varghese2024yolov8, he2017mask, li2019illumination, zhao2024detrs, lv2024rt}, which have been validated in industrial tasks. However, most of them mainly rely on single RGB input. In complex industrial environments, the performance of such single-modal frameworks often declines, owing to their diminished ability to extract discriminative object representations under non-ideal conditions.

\begin{figure}
    \centering
    \includegraphics[width=0.6\textwidth]{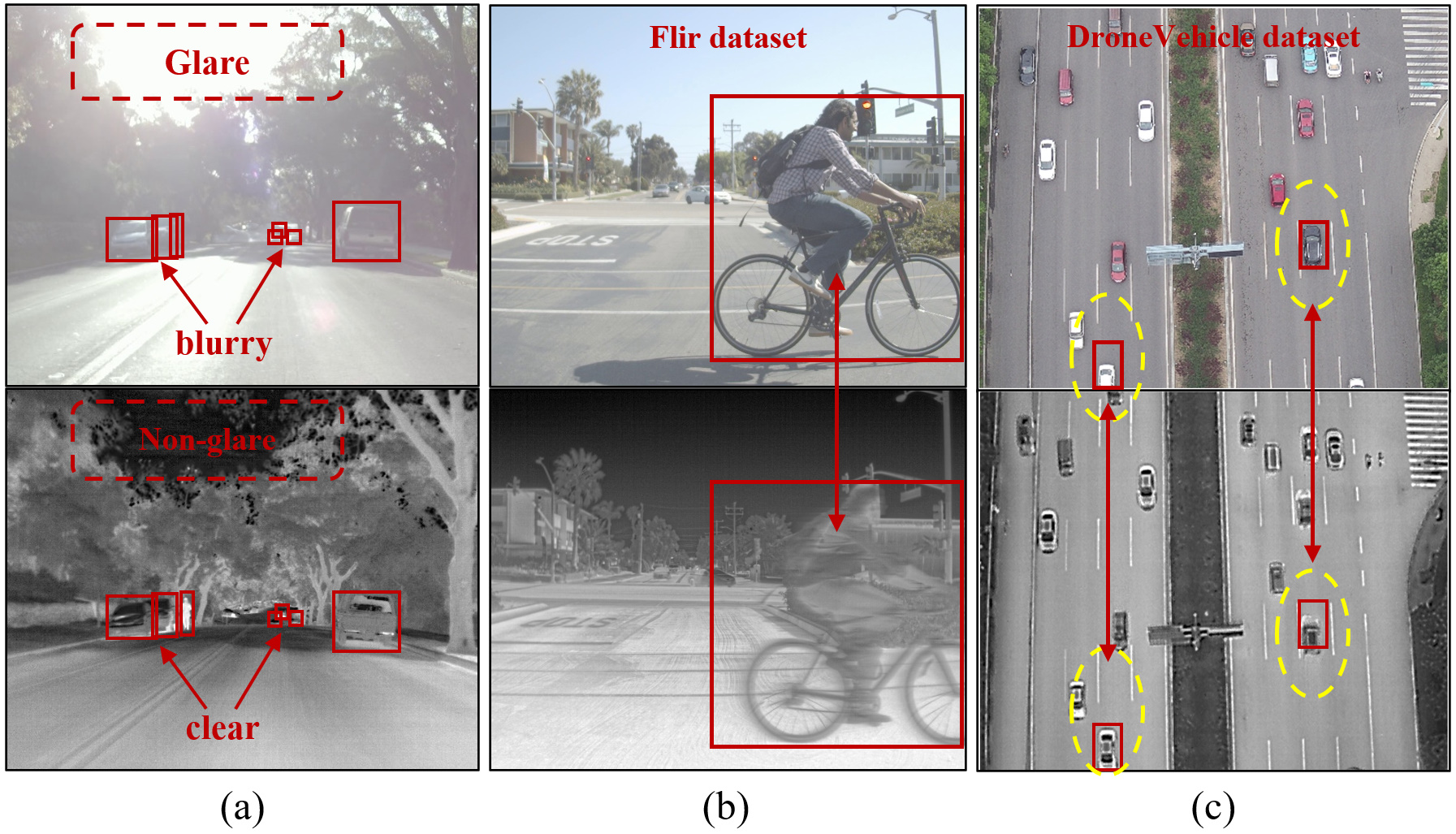}
    \caption{Multimodal characteristics and cross-modal misalignment. (a) Multimodal characteristics: Glare blurs the RGB, but does not affect the thermal. (b) Temporal capture misalignment. (c) Viewpoint-induced offset. The top row of the picture is an RGBs, and the bottom row is a thermals.}
    \label{fig1}
\end{figure}

Multimodal fusion is a key technique for addressing the issue of missing critical information in a single modality. It overcomes the limitations of a single sensor by fusing complementary data (e.g., thermal and depth). This integration mitigates local interference and missing details, enabling robust cross-modal validation. For example, thermal imaging is unaffected by glare and clearly distinguishes object contours through temperature differences, as shown in Fig.~\ref{fig1}(a). Furthermore, by sharing underlying features, multimodal approaches enrich global semantics, reduce reliance on any single modality, and significantly enhance the network's perceptual capabilities \cite{wang2023image}. 

While CNN- and ViT-based multimodal fusion techniques have demonstrated strong performance, several challenges remain. Firstly, CNNs are constrained by limited receptive fields, strong inductive biases, and difficulties in modeling long-range dependencies. Their weight-sharing mechanism across different input regions inherently restricts recognition accuracy and robustness \cite{pan2024open}. Transformer-based models (e.g., ViTs \cite{dosovitskiy2020image}, CrossViT \cite{chandrasiri2023cross}) offer global receptive fields but are limited to pairwise correlation modeling. Their quadratic computational complexity incurs a severe computational bottleneck when handling high-resolution images, thereby failing to meet the strict real-time constraints for industry. Secondly, while enhancing discriminative features, multimodal fusion inevitably introduces noise and redundancy, interfering with primary representation learning and increasing computational overhead. Moreover, due to differences in sensor frequencies and viewing angles, multimodal data often exhibits misalignment, see Fig.~\ref{fig1}(b) and (c). Such pixel-level offsets disrupt the global consistency of objects, which is crucial for understanding environmental context. Thus, effectively aligning modalities, designing efficient fusion mechanisms, and reducing computational overhead remain key research challenges.

Recently, Mamba-based models have gained widespread attention due to their powerful sequence modeling capabilities and linear computational complexity \cite{gu2024mamba}. Algorithms such as WaveMamba \cite{zhu2025wavemamba} and COMO \cite{liu2025cross} have demonstrated exceptional performance in multimodal detection. Despite high efficiency, Mamba-based approaches may encounter inherent structural challenges stemming from their reliance on linear scanning mechanisms. This flattening process potentially dilutes the inherent 2D spatial hierarchy of images, occasionally restricting the architecture primarily to sequential dependencies. This can partially constrain their efficacy in scenarios that demand meticulous structural and topological comprehension.

To address these challenges, we propose M2I2HA, a hypergraph-attention-based multimodal detection framework for RGB-IR object detection. Unlike conventional pairwise attention or sequential modeling, hypergraphs provide a flexible way to capture high-order relationships among multiple visual regions. In M2I2HA, the Intra-Hypergraph Enhancement module models high-order intra-modal dependencies with reduced parameter overhead, while the Inter-Hypergraph Fusion module exploits cross-hypergraph interactions to integrate complementary RGB-IR features. Furthermore, the M3FDFP block adaptively redistributes multi-level features to improve information flow and alleviate feature bottlenecks.

To the best of our knowledge, this is the first attempt to formulate cross-hypergraph interactions for multimodal object detection. Based on this formulation, we develop a lightweight hypergraph-attention framework that supports high-order cross-modal feature exchange under asymmetric feature lengths across modalities. Unlike pairwise attention or sequential modeling, M2I2HA provides an alternative mechanism for modeling many-to-many multimodal dependencies. In summary, the main contributions of this work are as follows:

(1) We propose M2I2HA, a hypergraph-attention-based framework for RGB-IR multimodal object detection. To the best of our knowledge, this is the first attempt to formulate cross-hypergraph interactions for multimodal detection, providing an alternative high-order interaction mechanism beyond pairwise attention and sequential modeling.

(2) We design an Inter-Hypergraph Fusion (IHF) module to establish high-order cross-modal hyperedge interactions between RGB and IR features. Unlike conventional fusion methods that mainly rely on pairwise feature aggregation or sequential modeling, IHF enables many-to-many complementary feature exchange across modalities and supports asymmetric feature lengths.

(3) We further develop a lightweight Intra-Hypergraph Enhancement (IHE) module and a Multimodal and Multi-level Feature Dynamic Fusion Pipeline (M3FDFP) block. The IHE module adapts HyperACE to RGB-IR multimodal detection through low-rank decomposition and sparse hyperedge assignment, reducing redundancy in intra-modal hyperedge modeling and improving parameter efficiency, while M3FDFP adaptively redistributes multi-level features to improve information flow.

(4) Extensive experiments on multiple public RGB-IR benchmarks demonstrate that M2I2HA achieves competitive detection performance with favorable computational efficiency. Ablation studies further verify the effectiveness of the proposed cross-hypergraph fusion and multi-level feature redistribution mechanisms.

\section{Related Works}
\subsection{Multimodal Fusion}
Recently, multimodal fusion has been widely applied in sensors, computer vision, and robotic perception. Early fusion (or data fusion) optimizes joint representations of heterogeneous data within the original space. By establishing cross-modal correlations or shared latent subspaces, it explicitly models feature interactions and achieves coordinated dimensionality reduction \cite{ramachandram2017deep}. Representative approaches include principal component analysis (PCA) \cite{he2010multimodal}, independent component analysis \cite{maglanoc2020multimodal}, canonical correlation analysis \cite{lei2016discriminative}, and maximum likelihood estimation \cite{ma2021maximum}. Although intuitive, this paradigm suffers from cross-modal spatiotemporal misalignment, noisy redundancy, and high resource costs. Recent advances in machine learning have propelled neural network-based methods to prominence due to their superior accuracy, speed, and generalization. Consequently, researchers are now actively investigating both feature-level and decision-level fusion strategies with neural networks. Feature fusion (or mid-level fusion), involves extracting high-level features from each modality, followed by fusing them. Common fusion strategies include feature concatenation \cite{jain2025informative}, tensor product \cite{zhang2025tensor}, and element-wise summation \cite{yuan2025feature}. It often incorporates skip connections to combine multi-scale features, which helps preserve details and is beneficial for the perception of small objects \cite{peng2025mlsa}. By introducing attention mechanisms, the networks can focus on key features, significantly enhancing perception accuracy \cite{yi2025hyfuser, guo2025cross}. For instance, Song et al. \cite{song2022cross} proposed a Cross-modal Contrastive Attention (CMCA) model to address the suboptimal results in automated medical report generation caused by data bias. R. Gnana Praveen et al. \cite{praveen2024recursive} introduced a Recursive Joint Cross-Modal Attention (RJCMA) mechanism to effectively capture intra- and inter-modal relationships across audio, visual, and textual for emotion recognition. Zhou et al. \cite{zhou2023cacfnet} presented a Cross-modal Attention Cascade Fusion Network (CACF-Net) for RGB-Turban scene analysis. Decision-level fusion, or late fusion, involves constructing separate feature extraction and decision networks for each modality independently. The results from each model are aggregated using various strategies, such as voting, maximum, minimum, or median. While this method eliminates the need for cross-modal feature synchronization and offers high flexibility, it inherently limits the interaction between data and modalities and remains susceptible to bias introduced by individual classifiers \cite{xu2025multi}.

\subsection{Multimodal Object Detection}
Single-modal detection often suffers from low accuracy in complex scenes. Multi-modal detection tackles this by integrating complementary data, offering an effective solution to performance degradation caused by illumination changes, occlusions, and multi-scale targets. SuperYOLO \cite{zhang2023superyolo} addressed the challenge of detecting multi-scale small objects in remote sensing imagery by extracting complementary information from multiple data sources, significantly boosting small object detection accuracy. Zhu et al. \cite{zhu2025wavemamba} introduced WaveMamba, which leverages the Discrete Wavelet Transform (DWT) to efficiently decompose and integrate complementary frequency-domain characteristics of RGB and infrared (IR) modalities, yielding significant detection gains. Zhou et al. \cite{zhou2025m} introduced M-SpecGene, a generalized foundation model for RGBT multispectral vision designed to learn modality-invariant representations from large-scale heterogeneous data via self-supervised pre-training. Wang et al. \cite{wang2025mff} proposed MFF-SDD, a multimodal framework leveraging bidirectional visual-textual cross-fusion and multiscale neighbor integration to identify tiny defects in BOPP films. Mahjourian et al. \cite{mahjourian2025multimodal} proposed a multimodal object detection method that integrates camera images and 3D sensor data within a modified Faster R-CNN framework, yielding substantial robustness and accuracy gains in industrial environments. COMO \cite{liu2025cross} extends VMamba with Mamba Interaction Blocks, Global-Local Scans, and Offset-Guided Fusion, jointly resolving modalities misalignment and enhancing detection accuracy of UAVs. 

\subsection{Hypergraph Attention Mechanism}
A hypergraph is an extension of an ordinary graph in graph theory, offering a more flexible framework for describing multi-way relationships. It has been widely applied in complex system modeling, such as in social networks \cite{meng2024link}, molecular interactions \cite{ye2024forecasting}, and recommendation systems \cite{sakong2024heterogeneous}. Qu et al. \cite{qu2021adaptive} proposed a multi-scale hypergraph-based network for efficient text detection. By employing a novel attention mechanism, this method enhances the semantic sensitivity of shallow features, effectively highlighting regions of interest while suppressing background interference. YOLOv13 \cite{lei2025yolov13} incorporates HyperACE, a hypergraph-based adaptive correlation enhancement mechanism. It dynamically mines latent high-order correlations transcending conventional pairwise associations, facilitating efficient cross-region and cross-scale feature fusion. Ferens et al. \cite{ferens2025hyperpose} introduced HyperPose, which integrates hypergraph networks into absolute camera pose regression. By generating input-adaptive weights for the regression head, it effectively bridges domain gaps and enhances robustness to diverse scene appearances.

\begin{figure*}
    \centering
    \includegraphics[width=\textwidth]{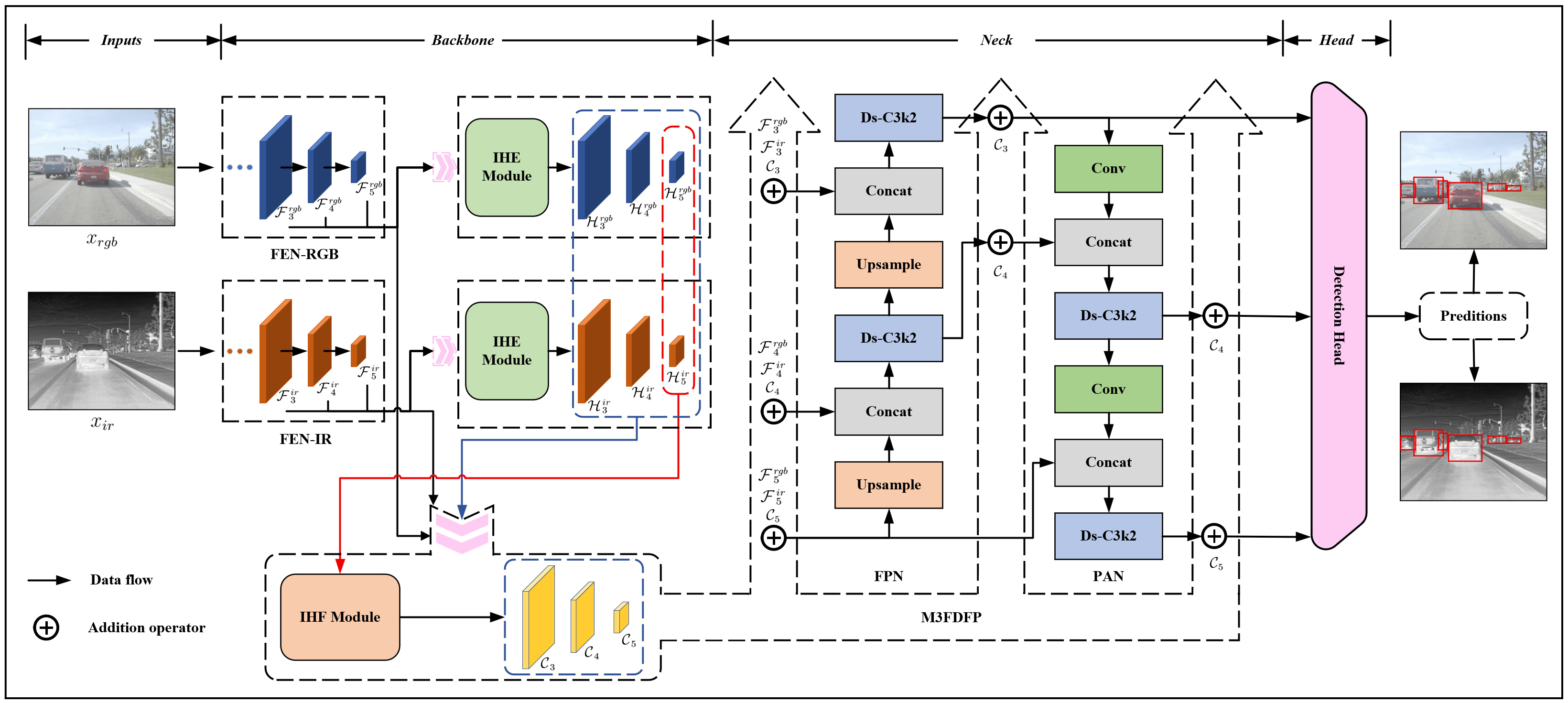}
    \caption{Overall architecture of M2I2HA. It adopts a YOLO-based architecture comprising a backbone, neck, and head. The backbone integrates two parallel feature extraction streams, accompanied by the IHE and IHF modules. They define a pipeline for multimodal feature extraction, intra-modal refinement, and cross-modal enhancement. The neck utilizes an FPN/PAN structure for multi-scale feature aggregation, while the head outputs final predictions. Additionally, the M3FDFP block is introduced to optimize feature flow and facilitate dynamic fusion across multiple stages.}
    \label{fig2}
\end{figure*}

\subsection{Preliminaries}
Graph structures excel at modeling pairwise relationships, but their descriptive capacity becomes limited when confronted with the multi-way associations prevalent in real-world scenarios. As a generalization of conventional graphs, hypergraphs allow hyperedges to connect any number of nodes, thereby offering an intuitive and flexible means to represent higher-order or group-wise relationships among entities.

A hypergraph can be formally represented as $\mathcal{H} = (\mathcal{V}, \mathcal{E})$, where $\mathcal{V} = \{v_1,v_2,...,v_n\}$ denotes a finite set of $n$ nodes, and $\mathcal{E} = \{e_1,e_2,...,e_m\}$ is a collection of $m$ hyperedges, each of which is a non empty subset of $\mathcal{V}$ (i.e., $e \subseteq  \mathcal{V}$ and $e \neq  \varnothing $). The structure of a hypergraph is commonly described by an incidence matrix $\mathbf{H}  \in \mathbb{R} ^{n\times m}$, whose elements are defined as follows:
\begin{equation}
\mathbf{H}_{i,j} =
\begin{cases}
1, & \text{if vertex } v_i \in \text{hyperedge } e_j \\
0, & \text{otherwise}
\end{cases}
\end{equation}

\subsection{Overall Architecture}
M2I2HA follows the YOLO framework and can be divided into three main components: the backbone, the neck, and the head, as illustrated in Fig. \ref{fig2}. The backbone consists of two CNN-based feature extraction networks (FENs), which perform multi-stride feature extraction for the RGB and another modality (taking thermal here) $ \{ x_{rgb}, x_{ir}\} $, respectively. The FENs share an identical structure and output multi-level features $ \{ \mathcal{F}^{rgb}_i, \mathcal{F}^{ir}_i\}_{i=3, 4, 5} $. Within each FEN, low-level layers are highly conducive to small object detection, while higher-level layers are robust to spatial misalignment and specialize in large objects, making them ideal for modeling cross-modal correlations. Assume the input sizes are $C\times W \times H$, where $C$, $W$, and $H$ denote the number of channels, width, and height, respectively. The detection output layers from P3 to P5 correspond to feature map sizes of $C_i \times (W/2^i) \times (H/2^i)$.

Within each FEN, Intra-Hypergraph Enhancement modules adaptively model high-order intra-modality correlations, producing enhanced features $\{ \mathcal{H}^{rgb}_i\}$ and $\{\mathcal{H}^{ir}_i \}$. High-level features $\{ \mathcal{H}^{rgb}_5, \mathcal{H}^{ir}_5 \}$ then feed into the Inter-Hypergraph Fusion module for cross-modal correlation modeling, yielding multi-scale representations $\{ \mathcal{C}_i \}$. Using high-level features here mitigates spatial misalignment and reduces computational overhead for real-time performance \cite{liu2025cross}. The M3FDFP block then integrates intra- and inter-modality enhanced features and dynamically distributes them across backbone-neck-head connections, preserving low-level details for small objects while enabling adaptive information flow. For the neck, M2I2HA adopts YOLOv13's design: large-kernel depthwise-separable convolutions (DSConv) form lightweight blocks, and a classic $\text{FPN}+\text{PAN}$ structure with bidirectional top-down and bottom-down pathways fuses shallow localization cues with deep semantics for multi-scale detection. Finally, the aggregated feature maps are sent to the head.

\begin{figure}
	\centering
    \includegraphics[width=0.5\textwidth]{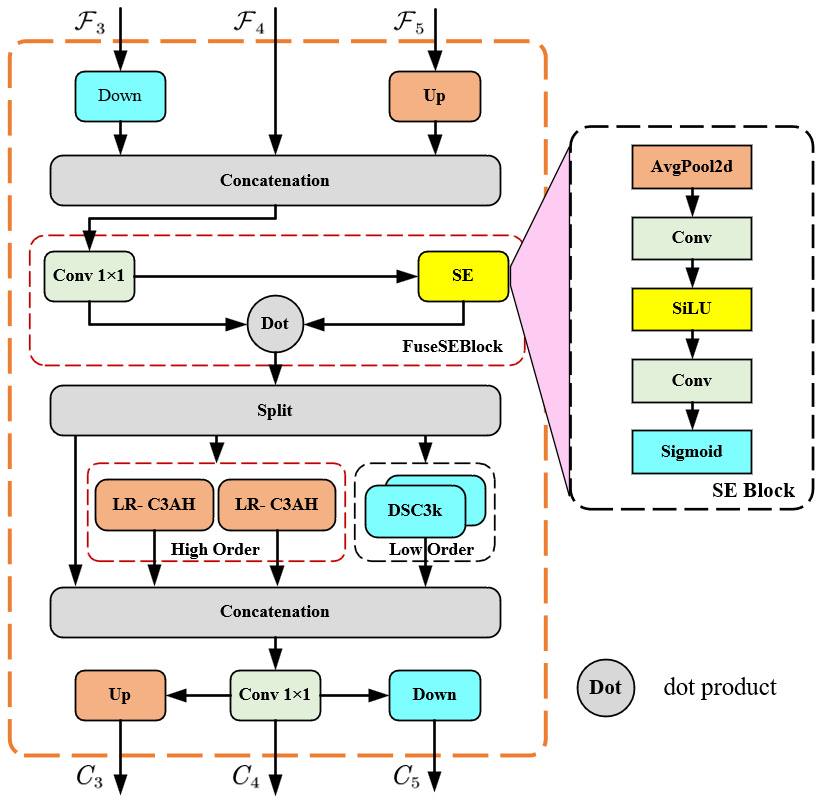}
	\caption{Architecture of the IHE module.}
	\label{fig3}
\end{figure}

\subsection{Intra-Hypergraph Enhancement Module}
Inspired by the Hypergraph-Based Adaptive Correlation Enhancement (HyperACE) module in YOLOv13, we develop the IHE module to capture high-order intra-modal relationships, as illustrated in Fig.~\ref{fig3}. Advancing beyond the original HyperACE, our design incorporates a dynamic fusion mechanism, low-rank decomposition, and sparsification techniques. This integration enhances representational capacity and robustness with minimal parameter and computational overhead. Here are the key components:

(1) \textbf{The FuseSEBlock}, which advances the fusion from static integration to dynamic calibration. Unlike HyperACE, which relies on naive feature concatenation and uniform convolution, we introduce a lightweight Squeeze-and-Excitation (SE) channel attention module, termed FuseSEBlock (see Fig.~\ref{fig3}). This design preserves multi-scale fusion capabilities while significantly enhancing the modeling of non-linear inter-channel dependencies. The calculation process is formulated as follows:
\begin{equation}
    \begin{cases}
        \mathcal{F}' = \text{Conv}\left( \text{Cat} \left( \text{Down} (\mathcal{F}_3), \, \mathcal{F}_4, \, \text{Up} (\mathcal{F}_5) \right)\right) \\
        \omega = \text{Sigmoid} \left( \text{Conv} \left( \text{SiLU} \left( \text{Conv} \left( \text{AvgPool} (\mathcal{F}') \right) \right) \right) \right) \\
        \mathcal{F} = \mathcal{F}' \bigodot \omega
    \end{cases}
\end{equation}
Where $\text{AvgPool}(\cdot)$ compresses spatial information. $\text{Conv}(\cdot)$ reduces computational complexity and enhances nonlinearity. $\text{SiLU}(\cdot)$ is the activation function. $\text{Sigmoid}(\cdot)$ normalizes the resulting weights. $\omega$ denotes the channel-wise attention weights, and $\bigodot$ represents element-wise multiplication with spatial broadcasting.

\begin{figure}
	\centering
    \includegraphics[width=0.6\textwidth]{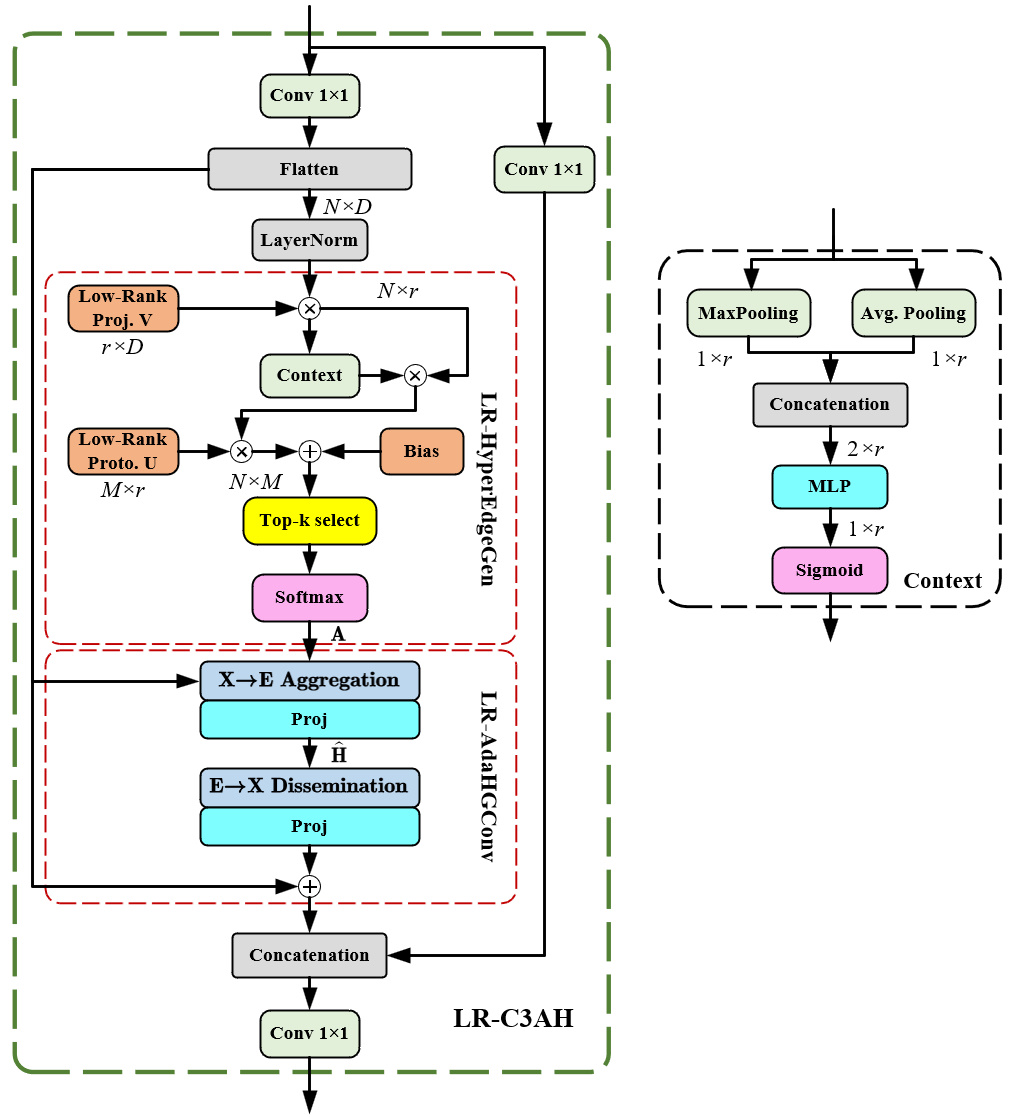}
	\caption{Architecture of the LR-C3AH block.}
	\label{fig4}
\end{figure}

(2) \textbf{The LR-C3AH Block} is the core improvement that distinguishes IHE from HyperACE, as illustrated in Fig.~\ref{fig4}. It integrates low-rank factorization and sparsification techniques into hypergraph computation to model high-order relationships, achieving high expressivity at a low computational cost. This strategy substantially alleviates parameter redundancy and computational complexity. 

\textbf{\textit{Hyperedge Generation:}}
Let the input feature map be $\mathcal{F}\in\mathbb{R}^{D\times H\times W}$, where the batch dimension is omitted for simplicity. After a $1\times1$ convolution, the feature map is flattened along the spatial dimensions and transposed to obtain the node feature matrix
\begin{equation}
    \mathbf{X}\in\mathbb{R}^{N\times D}, \quad N=H\times W
\end{equation}
Where $N$ denotes the number of nodes and $D$ denotes the node feature dimension.

In original HyperACE, a learnable base hyperedge prototype matrix $\mathbf{E}_{0}\in\mathbb{R}^{M\times D}$ is first constructed in the original $D$-dim feature space. Based on the global context extracted from the input nodes, a context mapping network generates a complete high-dimensional prototype offset $\Delta\mathbf{E}\in\mathbb{R}^{M\times D}$. The resulting input-dependent hyperedge prototypes are then given by
\begin{equation}
    \mathbf{E}=\mathbf{E}_{0}+\Delta\mathbf{E}
\end{equation}

The node features are linearly projected as
\begin{equation}
    \mathbf{Q}=\mathbf{X}\mathbf{W}_{q}, \quad \mathbf{W}_{q}\in\mathbb{R}^{D\times D}
\end{equation}

The node-hyperedge affinity scores are then computed in the original feature space
\begin{equation}
    \mathbf{S}_{\mathrm{H}} =\frac{\mathbf{Q}\mathbf{E}^{\mathrm{T}}}{\sqrt{D}}
\end{equation}
For clarity, the multi-head partition used in the implementation is omitted here, as it does not change the overall dimensionality or the asymptotic computational complexity. 

Finally, HyperACE normalizes the affinity scores along the node dimension
\begin{equation}
    \mathbf{A}_{\mathrm{H}}=\operatorname{Softmax}_{n}\left(\mathbf{S}_{\mathrm{H}}\right) \quad (n=1,...,N)
\end{equation}

The total computational complexity of HyperACE to generate the affinity matrix is about $\mathcal{O}(D^2(N+M)+NMD)$, which can be prohibitive for high-resolution images.

In contrast to HyperACE, which explicitly constructs complete hyperedge prototypes in the original feature space, we introduce a low-rank node projection matrix $\mathbf{V}\in\mathbb{R}^{r\times D}$, a low-rank hyperedge basis matrix $\mathbf{U}\in\mathbb{R}^{M\times r}$, and an edge-wise bias vector $\mathbf{b}\in\mathbb{R}^{M}$, where $r<D$ denotes the rank dimension.

The node features are first normalized and then projected into the $r$-dim low-rank space
\begin{equation}
    \mathbf{Z}=\operatorname{LayerNorm}(\mathbf{X})\mathbf{V}^{\mathrm{T}}\in\mathbb{R}^{N\times r}
    \label{eq:low_rank_node_projection}
\end{equation}
This operation avoids constructing node-hyperedge relationships directly in the original high-dimensional feature space.

HyperACE uses the input context to generate a complete $M\times D$ dynamic prototype offset. By contrast, the proposed method uses the context information only to adaptively recalibrate the low-rank dimensions. Specifically, global average pooling and global max pooling are applied to the low-rank node representations
\begin{equation}
    \mathbf{C}=\left[\operatorname{MeanPool}(\mathbf{Z});\operatorname{MaxPool}(\mathbf{Z})
    \right]\in\mathbb{R}^{1\times 2r}
\end{equation}

A lightweight multilayer perceptron is then used to generate a rank-wise context gate
\begin{equation}
    \mathbf{g}=2\cdot\sigma\left(\operatorname{MLP}(\mathbf{C})\right)\in\mathbb{R}^{1\times r}
    \label{eq:rank_gate}
\end{equation}
Where $\sigma(\cdot)$ denotes the sigmoid activation function. 

The context-enhanced low-rank node representations are obtained by rank-wise modulation
\begin{equation}
    \widetilde{\mathbf{Z}} = \mathbf{Z}\odot \mathbf{g}
    \label{eq:rank_modulation}
\end{equation}
Where $\odot$ denotes element-wise multiplication and $\mathbf{g}$ is broadcast across all nodes.

The node-hyperedge affinity score matrix is subsequently computed directly in the low-rank space
\begin{equation}
    \mathbf{S}=\frac{\widetilde{\mathbf{Z}}\mathbf{U}^{\mathrm{T}}}{\sqrt{r}}+\mathbf{b}\in\mathbb{R}^{N\times M}
    \label{eq:low_rank_affinity}
\end{equation}
Importantly, Eq.~\eqref{eq:low_rank_affinity} directly matches the low-rank node representations with the low-rank hyperedge bases and does not reconstruct an intermediate $M\times D$ hyperedge prototype matrix.

Based on the affinity scores in Eq.~\eqref{eq:low_rank_affinity}, a Top-$k$ selection strategy is employed to retain only the most relevant hyperedges. This produces a sparse node--hyperedge participation pattern and avoids propagating messages through weakly related hyperedges.

For the node-wise routing mode, the $K$ hyperedges with the highest affinity scores are independently selected for each node
\begin{equation}
    \left(
        \widetilde{\mathbf{S}}_{i,:},
        \mathbf{I}_{i,:}
    \right)
    =
    \operatorname{TopK}_{K}
    \left(
        \mathbf{S}_{i,:}
    \right)
    \quad
    (i=1,\ldots,N)
    \label{eq:node_topk}
\end{equation}
Where $\mathbf{I}_{i,:}\in\mathbb{N}^{K}$ contains the indices of the selected hyperedges and $\widetilde{\mathbf{S}}_{i,:}\in\mathbb{R}^{K}$ contains the corresponding retained scores. 

The sparse participation weights are then obtained by normalizing the retained scores over the selected hyperedges
\begin{equation}
    \mathbf{A}_{i,:}
    =
    \operatorname{Softmax}_{k}
    \left(
        \widetilde{\mathbf{S}}_{i,:}
    \right) \quad (k=1,...,K)
    \label{eq:sparse_participation}
\end{equation}
Accordingly, $\mathbf{A}\in\mathbb{R}^{N\times K}$ represents the final sparse node-hyperedge participation matrix.

In addition to node-wise routing, we introduce a global routing mode for more efficient hyperedge selection. In this mode, the context-enhanced low-rank features are first globally aggregated
\begin{equation}
    \mathbf{z}_{g}
    =
    \operatorname{MeanPool}
    \left(
        \widetilde{\mathbf{Z}}
    \right)
    \in
    \mathbb{R}^{1\times r}
    \label{eq:global_mean_pool}
\end{equation}

Global-level routing scores are then computed to select a shared subset of $K$ hyperedges
\begin{equation}
    \mathbf{s}_{g}
    =
    \frac{
        \mathbf{z}_{g}
        \mathbf{U}^{\mathrm{T}}
    }{
        \sqrt{r}
    }
    +
    \mathbf{b}
    \label{eq:global_routing_scores}
\end{equation}

\begin{equation}
    \mathbf{I}
    =
    \operatorname{TopK}_{k}
    \left(
        \mathbf{s}_{g}
    \right)
    \label{eq:global_selected_hyperedges}
\end{equation}
All nodes within the same sample share the selected hyperedge subset $\mathbf{I}$. Node-hyperedge affinity scores are subsequently evaluated only with respect to the selected low-rank hyperedge bases, and a softmax operation is applied over the resulting $K$ candidates.

The node-wise mode allows different nodes to establish individual hyperedge connections and therefore provides greater representational flexibility. In comparison, the global mode shares the selected hyperedges among all nodes within a sample, enabling more compact matrix multiplication and lower computational overhead. 

The total computational complexity of IHE for generating the affinity matrix in global mode and node-wise mode is approximately $\mathcal{O}(BNr(D+K))$ and $\mathcal{O}(BNr(D+M))$, respectively. They are significantly lower than that of HyperACE as $r\ll D$.

\textbf{\textit{Node $\rightarrow$ Hyperedge Aggregation:}}
The sparse node-hyperedge participation weights are obtained from the preceding LR-HyperEdgeGen block. For node $\mathbf{x}_i$, let $\mathcal{I}_i$ denote the indices of the $K$ selected hyperedges. The participation weight between node $\mathbf{x}_i$ and hyperedge $j$ is defined as
\begin{equation}
    A_{i,j}
    =
    \begin{cases}
        \displaystyle
        \frac{\exp(S_{i,j})}
        {\sum_{\ell\in\mathcal{I}_i}\exp(S_{i,\ell})}
        & j\in\mathcal{I}_i \\[8pt]
        0
        & j\notin\mathcal{I}_i
    \end{cases}
    \label{eq:sparse_participation_weight}
\end{equation}
Where $\mathbf{S}$ denotes the node-hyperedge affinity score matrix produced in the low-rank space. Consequently,
\begin{equation}
    \sum_{j\in\mathcal{I}_i} A_{i,j}=1
\end{equation}
holds for each node. In the node-wise routing mode, different nodes may select different hyperedge subsets, whereas in the global routing mode, all nodes within the same sample share an identical selected subset, i.e., $\mathcal{I}_i=\mathcal{I}$.

Different from HyperACE, before hypergraph aggregation, each node feature is projected into a compressed message space
\begin{equation}
    \mathbf{m}_i
    =
    \phi_{\mathrm{in}}(\mathbf{x}_i)
    =
    \mathbf{x}_i\mathbf{W}_{\mathrm{in}}
    \in\mathbb{R}^{P}
    \label{eq:input_message_projection}
\end{equation}
Where $\mathbf{W}_{\mathrm{in}}\in\mathbb{R}^{D\times P}$ and $P<D$ denotes the message dimension.

For hyperedge $\mathbf{e}_j$, its weighted degree is computed as
\begin{equation}
    d_j
    =
    \sum_{i\in\mathcal{N}(j)} A_{i,j}
    \label{eq:hyperedge_degree}
\end{equation}
Where $\mathcal{N}(j)=\left\{i \mid j \in \mathcal{I}_i\right\}$ denotes the set of nodes connected to hyperedge $j$.

The feature of hyperedge $\mathbf{e}_j$ is then obtained through degree-normalized node-to-hyperedge aggregation
\begin{equation}
    \mathbf{H}_j
    =
    \frac{1}{d_j+\varepsilon}
    \sum_{i\in\mathcal{N}(j)}
    A_{i,j}\mathbf{m}_i
    \label{eq:node_to_hyperedge}
\end{equation}
Where $\varepsilon$ is a small constant introduced for numerical stability. The aggregated hyperedge feature is subsequently refined by a nonlinear projection function:
\begin{equation}
    \widehat{\mathbf{H}}_j
    =
    \rho_{\mathrm{edge}}(\mathbf{H}_j)
    \label{eq:hyperedge_projection}
\end{equation}
Where $\rho_{\mathrm{edge}}(\cdot)$ consists of a linear projection, GELU activation, dropout, and layer normalization.

\textbf{\textit{Hyperedge $\rightarrow$ Node Dissemination:}}
After the hyperedge representations have been updated, the information carried by the selected hyperedges is disseminated back to their associated nodes. For node $\mathbf{x}_i$, the received hyperedge message is computed as
\begin{equation}
    \mathbf{y}_i
    =
    \sum_{j\in\mathcal{I}_i}
    A_{i,j}\widehat{\mathbf{H}}_j
    \in\mathbb{R}^{P}
    \label{eq:hyperedge_to_node}
\end{equation}

The resulting message is then projected from the compressed message space back to the original feature dimension
\begin{equation}
    \Delta\mathbf{x}_i=\rho_{\mathrm{node}}(\mathbf{y}_i)\in\mathbb{R}^{D}
    \label{eq:node_output_projection}
\end{equation}
Where $\rho_{\mathrm{node}}(\cdot)$ denotes the output projection from the $P$-dimensional message space to the original $D$-dim node feature space.

Finally, the node feature is updated through a LayerScale residual connection
\begin{equation}
    \widehat{\mathbf{x}}_i
    =
    \mathbf{x}_i
    +
    \boldsymbol{\gamma}
    \odot
    \operatorname{Dropout}
    \left(
        \Delta\mathbf{x}_i
    \right)
    \label{eq:node_residual_update}
\end{equation}
Where $\boldsymbol{\gamma}\in\mathbb{R}^{D}$ is a learnable channel-wise scaling parameter and $\odot$ denotes element-wise multiplication. This residual formulation preserves the original node representation while adaptively incorporating high-order information propagated through the selected hyperedges.

\subsection{Inter-Hypergraph Fusion Module}
\begin{figure}
    \centering
    \includegraphics[width=0.6\textwidth]{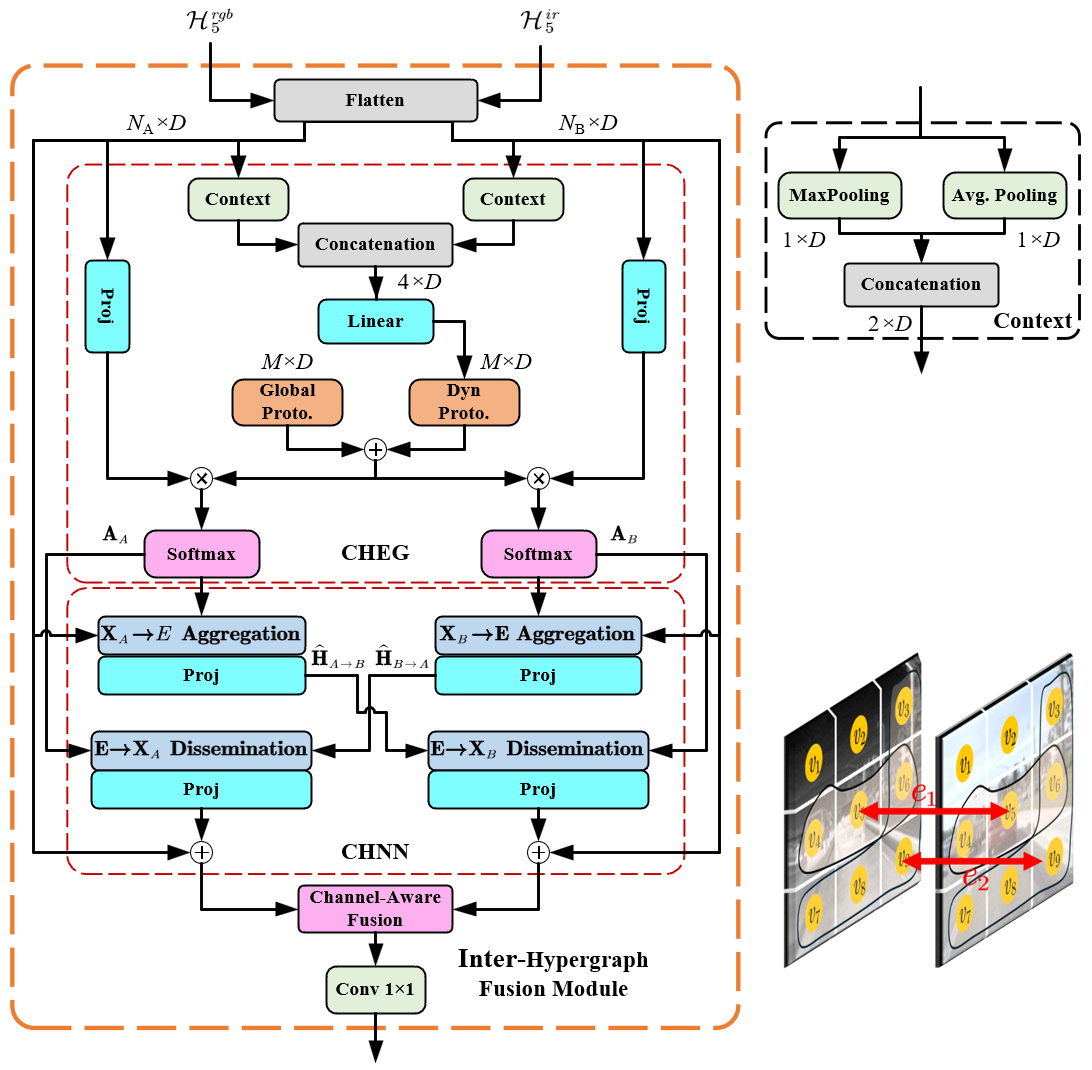}
	\caption{Illustration of IHF module with hyperedge across modalities.}
    \label{fig5}
\end{figure}

To address the challenges of modeling cross-modal correlations, we propose the IHF module, as illustrated in Fig. \ref{fig5}. It consists of three blocks: CrossHyperEdgeGen block (CHEG), CrossHyperConv block (CHNN), and GateFusion block. 

The enhanced features from the IHE module $\{ \mathcal{H}^{rgb}_5, \mathcal{H}^{ir}_5 \}$ are first flattened and transposed to a shape of $N_A(N_B) \times D$ and then fed into the CHEG block to generate the sparse node-hyperedge participation matrix $\mathbf{A}_A$ and $\mathbf{A}_B$. Following this, $\mathbf{A}_A$ and $\mathbf{A}_B$, along with their corresponding features, are fed into the CHNN block for cross-modal enhanced feature generation. The enhanced features pass through a Channel-Aware Fusion block to adaptively fuse modalities

\begin{equation}
\begin{cases}
    \tilde{\mathcal{H}}_{5}^{\mathrm{rgb}},\ \tilde{\mathcal{H}}_{5}^{\mathrm{ir}} = \operatorname{Flatten}\left(\mathcal{H}_{5}^{\mathrm{rgb}},\ \mathcal{H}_{5}^{\mathrm{ir}}\right) \\[6pt]
    \mathbf{A}_A,\ \mathbf{A}_B = \operatorname{CHEG}\left(\tilde{\mathcal{H}}_{5}^{\mathrm{rgb}},\ \tilde{\mathcal{H}}_{5}^{\mathrm{ir}}\right) \\[6pt]
    \tilde{\mathcal{H}}_{e}^{\mathrm{rgb}},\ \tilde{\mathcal{H}}_{e}^{\mathrm{ir}} = \operatorname{CHNN}\left(\tilde{\mathcal{H}}_{5}^{\mathrm{rgb}},\ \tilde{\mathcal{H}}_{5}^{\mathrm{ir}},\ \mathbf{A}_A,\ \mathbf{A}_B\right) \\[6pt]
    \mathbf{Y} = \operatorname{Conv}\Bigl(\operatorname{ChanAwareFusion}\left(\tilde{\mathcal{H}}_{e}^{\mathrm{rgb}},\ \tilde{\mathcal{H}}_{e}^{\mathrm{ir}}\right)\Bigr)
\end{cases}
\end{equation}

\textbf{\textit{Hyperedge Generation:}}
The IHF module leverages our newly formulated cross-hypergraph theory to execute the reciprocal cross-update of hyperedge and node matrices. Suppose there is a set of hyperedges $\mathbf{E} \in \mathbb{R}^{M \times D}$ simultaneously connecting a group of nodes $\mathbf{X}_A \in \mathbb{R}^{N_A \times D}$ and another group of nodes $\mathbf{X}_B \in \mathbb{R}^{N_B \times D}$, as shown in Fig. \ref{fig5}. 

The node features are first normalized and then projected as
\begin{equation}
    \begin{cases}
    \widetilde{\mathbf{X}}_A=\mathbf{X}_A\mathbf{W}_A^{\mathrm{T}} \in\mathbb{R}^{N_A\times D}\\
    \widetilde{\mathbf{X}}_B=\mathbf{X}_B\mathbf{W}_B^{\mathrm{T}} \in\mathbb{R}^{N_B\times D}
    \end{cases}
\end{equation}
Where, $\mathbf{W}_A, \mathbf{W}_B \in \mathbb{R}^{D\times D}$ are learnable projection matrices.

Subsequently, a learnable hyperedge prototype base matrix $\mathbf{E}_0 \in \mathbb{R}^{M \times D}$ is built. The contextual features of the two nodes are extracted separately and then combined into a unified contextual feature vector.
\begin{equation}
    \begin{cases}
    \mathbf{c}_A=\left[\operatorname{MeanPool}(\mathbf{X}_A);\operatorname{MaxPool}(\mathbf{X}_A) \right] \in\mathbb{R}^{2\times D}\\
    \mathbf{c}_B=\left[\operatorname{MeanPool}(\mathbf{X}_B);\operatorname{MaxPool}(\mathbf{X}_B) \right] \in\mathbb{R}^{2\times D}\\
    \mathbf{C}=[\mathbf{c}_A;\mathbf{c}_B] \in\mathbb{R}^{4\times D}
    \end{cases}
\end{equation}

Then, the prototype bias $\Delta \mathbf{E} \in \mathbb{R}^{M \times D}$ is obtained via a linear transformation of the combined contextual features. This bias enhances the model's ability to discriminate and represent specific data patterns. Final hyperedge prototype matrix is then constructed by adding the bias to the learnable hyperedge prototype base matrix. 
\begin{equation}
    \begin{cases}
    \Delta \mathbf{E} = \operatorname{Linear}(\mathbf{C}) \\
    \mathbf{E} = \mathbf{E}_0 + \Delta \mathbf{E}
    \end{cases}
\end{equation}

The node-hyperedge affinity score matrixs are subsequently computed as
\begin{equation}
    \begin{cases}
    \mathbf{S}_A=\frac{\widetilde{\mathbf{X}}_A\mathbf{E}^{\mathrm{T}}}{\sqrt{D}}\in\mathbb{R}^{N_A\times M}\\
    \mathbf{S}_B=\frac{\widetilde{\mathbf{X}}_B\mathbf{E}^{\mathrm{T}}}{\sqrt{D}}\in\mathbb{R}^{N_B\times M}
    \end{cases}
\end{equation}

$\mathbf{S}_A$ and $\mathbf{S}_B$ are used to calculate the sparse node-hyperedge participation matrix $\mathbf{A}_A$ and $\mathbf{A}_B$ with reference to Eqs.~\eqref{eq:node_topk}-\eqref{eq:global_selected_hyperedges}. Notably, we do not employ low-rank decomposition during hypergraph fusion, as we found that it significantly impairs model performance. Please refer to the experimental section for a detailed discussion.

Based on the participation matrices $\mathbf{A}_{A}\in\mathbb{R}^{N_A\times K}$ and $\mathbf{A}_{B}\in\mathbb{R}^{N_B\times K}$ generated by the CHEG block, cross-modal hypergraph message propagation is performed by the CHNN block. 

\textbf{\textit{Node $\rightarrow$ Hyperedge Aggregation:}}
The degree matrices of the selected hyperedges are calculated as
\begin{equation}
\begin{aligned}
    \mathbf{D}_{A}
    &=
    \operatorname{Diag}
    \left(
        \mathbf{A}_{A}^{\mathrm{T}}
        \mathbf{1}_{N_A}
    \right)\\
    \mathbf{D}_{B}
    &=
    \operatorname{Diag}
    \left(
        \mathbf{A}_{B}^{\mathrm{T}}
        \mathbf{1}_{N_B}
    \right)
\end{aligned}
\label{eq:ihf_hyperedge_degree}
\end{equation}
Where $\mathbf{1}_{N} \in \mathbb{R}^{N}$ is a vector of ones.

The modality-specific hyperedge representations are then obtained through degree-normalized node-to-hyperedge aggregation:
\begin{equation}
\begin{aligned}
    \mathbf{H}_{A}
    &=
    \mathbf{D}_{A}^{-1}
    \mathbf{A}_{A}^{\mathrm{T}}
    \mathbf{X}_{A},\\
    \mathbf{H}_{B}
    &=
    \mathbf{D}_{B}^{-1}
    \mathbf{A}_{B}^{\mathrm{T}}
    \mathbf{X}_{B}.
\end{aligned}
\label{eq:ihf_node_to_hyperedge}
\end{equation}
Here, $\mathbf{H}_{A},\mathbf{H}_{B}\in\mathbb{R}^{K\times P}$ represent the hyperedge features aggregated from modalities $A$ and $B$, respectively. The degree normalization alleviates feature-scale variations caused by different node numbers and connection densities.

\textbf{\textit{Cross-Modal Hyperedge $\rightarrow$ Node Dissemination:}}
Cross-modal interaction is conducted at the hyperedge level. The hyperedge features aggregated from one modality are first transformed into the space of the other modality as
\begin{equation}
\begin{aligned}
    \widehat{\mathbf{H}}_{B\rightarrow A}
    &=
    \rho_{B\rightarrow A}
    \left(
        \mathbf{H}_{B}
    \right)\\
    \widehat{\mathbf{H}}_{A\rightarrow B}
    &=
    \rho_{A\rightarrow B}
    \left(
        \mathbf{H}_{A}
    \right)
\end{aligned}
\label{eq:ihf_cross_hyperedge_projection}
\end{equation}
Where $\rho_{B\rightarrow A}(\cdot)$ and $\rho_{A\rightarrow B}(\cdot)$ denote nonlinear cross-modal hyperedge projection functions.

Each modality subsequently retrieves the hyperedge information from the opposite modality using its own sparse participation matrix
\begin{equation}
\begin{aligned}
    \mathbf{R}_{A}
    &=
    \mathbf{A}_{A}
    \widehat{\mathbf{H}}_{B\rightarrow A}
    \in
    \mathbb{R}^{N_A\times D}\\
    \mathbf{R}_{B}
    &=
    \mathbf{A}_{B}
    \widehat{\mathbf{H}}_{A\rightarrow B}
    \in
    \mathbb{R}^{N_B\times D}
\end{aligned}
\label{eq:ihf_hyperedge_to_node}
\end{equation}

The received cross-modal messages are then projected as
\begin{equation}
\begin{aligned}
    \Delta\mathbf{X}_{A}
    &=
    \mathbf{R}_{A}
    \mathbf{W}_{A}^{\mathrm{out}}\\
    \Delta\mathbf{X}_{B}
    &=
    \mathbf{R}_{B}
    \mathbf{W}_{B}^{\mathrm{out}}
\end{aligned}
\label{eq:ihf_output_projection}
\end{equation}
Where $\mathbf{W}_{A}^{\mathrm{out}}, \mathbf{W}_{B}^{\mathrm{out}}\in\mathbb{R}^{D\times D}$.

Finally, gated LayerScale residual connections are employed to incorporate the received cross-modal information
\begin{equation}
\begin{aligned}
    \widehat{\mathbf{X}}_{A}
    &=
    \mathbf{X}_{A}
    +
    \sigma
    \left(
        \boldsymbol{\alpha}_{A}
    \right)
    \odot
    \boldsymbol{\gamma}_{A}
    \odot
    \operatorname{Dropout}
    \left(
        \Delta\mathbf{X}_{A}
    \right)\\
    \widehat{\mathbf{X}}_{B}
    &=
    \mathbf{X}_{B}
    +
    \sigma
    \left(
        \boldsymbol{\alpha}_{B}
    \right)
    \odot
    \boldsymbol{\gamma}_{B}
    \odot
    \operatorname{Dropout}
    \left(
        \Delta\mathbf{X}_{B}
    \right)
\end{aligned}
\label{eq:ihf_node_update}
\end{equation}
Where $\boldsymbol{\alpha}_{A}$ and $\boldsymbol{\alpha}_{B}$ are learnable channel-wise gates, while $\boldsymbol{\gamma}_{A},\boldsymbol{\gamma}_{B}\in\mathbb{R}^{D}$ denote the LayerScale parameters. This formulation preserves the original modality-specific node representations while adaptively incorporating high-order information propagated from the opposite modality.

\subsection{M3FDFP Block}
We develop the Multimodal and Multi-level Feature Dynamic Fusion Pipeline (M3FDFP) to facilitate dynamic feature distribution. Motivated by the Data Processing Inequality (DPI) \cite{sinha2020d2rl}, which implies that information inevitably decays through successive processing layers (i.e., $\text{MI}(S_1, S_{n-1}) \geq \text{MI}(S_1, S_{n})$), we adopt dense connectivity strategies \cite{huang2017densely} to mitigate information loss.

M3FDFP aggregates three feature sets: backbone features, intra-modality enhanced features, and inter-modality fused features. These are redistributed to the neck pipeline via an adaptive fusion mechanism (see Fig. \ref{fig2}). Formally, the fused feature map $\widetilde{\mathcal{F}}_i$ is defined as:
\begin{equation}
    \widetilde{\mathcal{F}}_i = \text{ModalFuseSE}(\mathcal{F}^{rgb}_i,\mathcal{F}^{ir}_i)+\alpha \mathcal{H}^{rgb}_i+\beta \mathcal{H}^{ir}_i+\gamma \mathcal{C}_i
\end{equation}
Where $\text{ModalFuseSE}(\cdot)$ is a channel attention fusion module structurally similar to the FuseSEBlock. $\alpha, \beta, \gamma$ are learnable scalars that dynamically weight the contributions of enhanced features. This design effectively injects correlation information into various pipeline stages, significantly enhancing scene perception.

\section{Experiments}
\subsection{Datasets and Implementation Details}
To thoroughly evaluate the proposed M2I2HA framework, extensive experiments were conducted on four benchmark datasets: DroneVehicle \cite{sun2022drone}, FLIR-Aligned \cite{zhang2020multispectral}, LLVIP \cite{jia2021llvip}, and VEDAI \cite{razakarivony2016vehicle}. 

State-of-the-art representative methods were selected for comparison, including: SuperYOLO \cite{zhang2023superyolo}, ICAFusion \cite{shen2024icafusion}, CFT \cite{qingyun2021cross}, GHOST \cite{zhang2023guided}, GM-DETR \cite{xiao2024gm}, COMO \cite{liu2025cross}, and WaveMamba \cite{zhu2025wavemamba}.

All experiments were conducted on a Linux server using PyTorch, equipped with an AMD Ryzen 9 9950X CPU and an NVIDIA A100 GPU (80GB). To ensure fair comparison, all models were trained from scratch using the SGD optimizer with identical hyperparameters. Input images were resized to $640 \times 640$. During training, standard augmentations (e.g., random flip, Mosaic) were applied. Inference was performed in FP32 mode without augmentation. Computational efficiency is reported as Frames Per Second (FPS), calculated as the average end-to-end latency including data loading, forward propagation, and post-processing. The detailed and inference configurations are summarized in Table \ref{table0}.

\begin{table}[htbp]
    \centering
    \caption{Detailed training and inference configurations.}
    \resizebox{0.5\textwidth}{!}{
        \begin{tabular}{@{}l|c@{}}
            \toprule
            \textbf{Items} & \textbf{Values} \\
            \midrule
            \multicolumn{2}{c}{\textbf{Training Settings}} \\
            \midrule
            Image size & $640 \times 640$ \\
            Batch size & $64$ \\
            AMP & True \\
            \midrule
            \multicolumn{2}{c}{\textbf{Optimization}} \\
            \midrule
            Optimizer & SGD \\
            Initial LR & $0.01$ \\
            Final LR factor & $0.01$ \\
            Weight decay & $5 \times 10^{-4}$ \\
            Momentum & $0.937$ \\
            Warm-up epochs & $3$ \\
            \midrule
            \multicolumn{2}{c}{\textbf{Data Augmentation}} \\
            \midrule
            Mosaic & $1.0$ \\
            MixUp & $0.0$ \\
            Copy-paste & $0.0$ \\
            HSV-H & $0.015$ \\
            HSV-S & $0.7$ \\
            HSV-V & $0.4$ \\
            \midrule
            \multicolumn{2}{c}{\textbf{Loss Weights}} \\
            \midrule
            BOX loss & $7.5$ \\
            CLS loss & $0.5$ \\
            DFL loss & $1.5$ \\
            \midrule
            \multicolumn{2}{c}{\textbf{Post-processing}} \\
            \midrule
            NMS IoU threshold & $0.7$ \\
            Conf threshold (validation) & $0.001$ \\
            Conf threshold (inference) & $0.25$ \\
            \bottomrule
        \end{tabular}%
    }
    \label{table0}
\end{table}

To thoroughly evaluate the detection performance of different methods, the widely adopted mean Average Precision was employed as the primary quantitative metric. Three main variants of mAP with different levels of strictness were reported.

\subsection{Evaluation on the DroneVehicle Dataset}
DroneVehicle is a large-scale dual-modal benchmark comprising 28439 RGB-T pairs captured at 40-130m altitudes under diverse conditions. It poses severe challenges, including dramatic scale variations, dense small objects, occlusions, and illumination changes.

\begin{figure}
    \centering
    \includegraphics[width=0.8\textwidth]{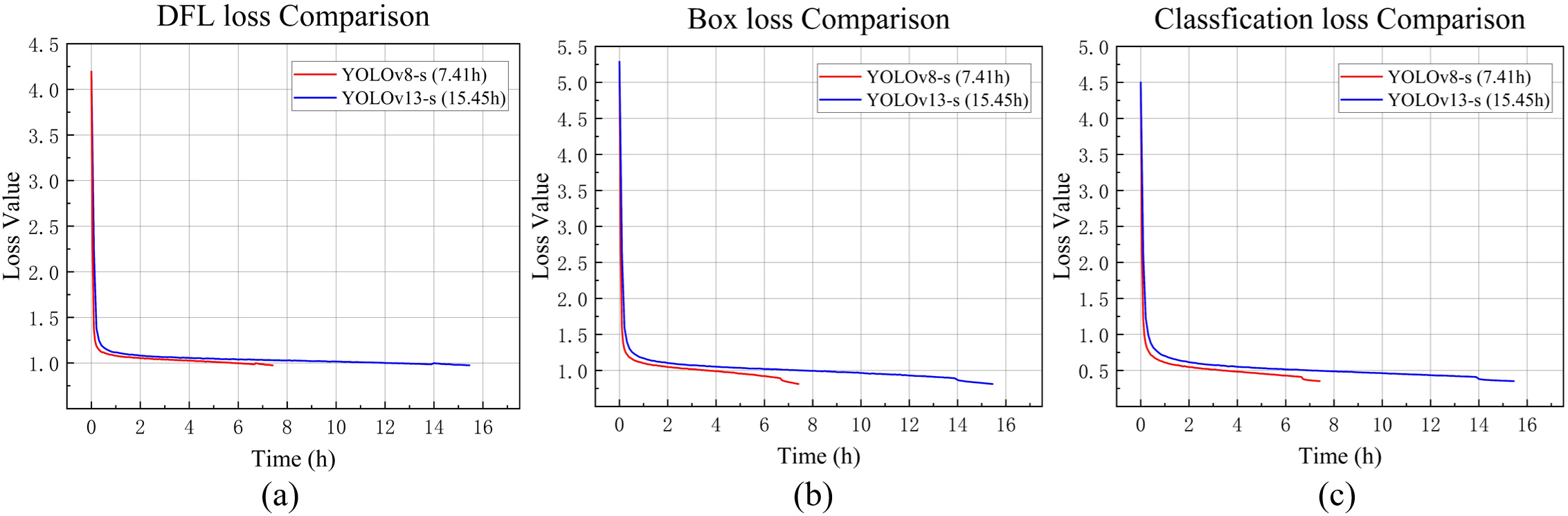}
    \caption{The loss curves of our YOLOv8s-based and YOLOv13s-based models.}
    \label{fig6}
\end{figure}

Table \ref{table1} summarizes the quantitative results. To ensure fairness, all models were trained from scratch for 150 epochs. Our YOLOv8s-based model achieves the best performance, attaining 85.4\% mAP@.5, 75.7\% mAP@.75, and 63.4\% mAP. The lightweight YOLOv8n variant secures the second-best 84.2\% mAP@.5, validating that our cross-modal enhancement module effectively compensates for lightweight networks' limited capacity. Conversely, YOLOv13-backbone models underperformed and exhibited slower convergence (Fig. \ref{fig6}). This is attributed to their extensive use of Depthwise Separable Convolutions; while parameter-efficient, DSC introduces greater optimization challenges when training from scratch than standard convolutions, hindering feature representation. Among other comparative methods, COMO ranks second in 75.1\% mAP@.75 and 63.3\% mAP by leveraging a Mamba Interaction Block for long-range dependencies and an Offset-guided Fusion to mitigate high-level spatial misalignment. CFT and WaveMamba also show competitive results.
\begin{table}[htbp]
	\caption{Experiment results on the DroneVehicle dataset.}
    \centering
    \resizebox{\textwidth}{!}{
		\begin{tabular}[htbp]{@{}c|c|c|ccccc|ccc@{}}
			\hline
			Methods           & Modality & Backbone  & Car & Truck & Bus  & Van  & Freight Car & mAP@.5 (\%)   & mAP@.75 (\%)   & mAP (\%)   \\
			\hline
			YOLOv8 (ADICS'2024)  & RGB      & YOLOv8s   & 92.1 &  54.7 & 67.9 & 93.0 &    56.4     &     72.8    &     49.4     & 44.8              \\
			YOLOv8 (ADICS'2024)  & IR       & YOLOv8s   & 97.8 &  65.9 & 72.5 & 95.2 &    60.8     &     78.5    &     68.5     & 57.1              \\
			YOLOv13 (arXiv'2025) & RGB      & YOLOv13s  & 91.8 &  52.2 & 65.0 & 92.0 &    54.3     &     71.1    &     48.7     & 44.0              \\
			YOLOv13 (arXiv'2025) & IR       & YOLOv13s  & 97.9 &  67.9 & 73.6 & 96.4 &    62.3     &     79.6    &     68.8     & 57.3              \\
            \hline
			CFT (arXiv'2022)     & RGB+IR   & YOLOv5s   &\textbf{98.5} &74.6  & 79.6 & \underline{96.9} & \underline{69.7} & 83.9 &  73.5  & 61.3   \\
            SuperYOLO (TGRS'2023)& RGB+IR   & YOLOv5s   & 97.8 &  \textbf{78.9} & 66.5 & 96.5 &    67.4     &     81.4    &     70.1     & 58.6     \\
            GHOST (TGRS'2023)    & RGB+IR   & YOLOv5s   & 97.2 &  75.6 & 70.9 & 95.8 &    66.5     &     81.2    &     72.7     & 59.1     \\
            ICAFusion (PR'2024)  & RGB+IR   & YOLOv5s   & 98.2 &  72.1 & 79.5 & 96.5 &    67.1     &     82.7    &     72.7     & 60.5     \\
            GM-DETR (CVPR'2024)  & RGB+IR   & RT-DETR   & 92.3 &  74.9 & \underline{81.0} & 91.2 &    65.7     &     81.0    &     70.9     & 57.8     \\
            COMO (IF'2025)       & RGB+IR   & YOLOv8s   & 98.1 &  75.8 & 80.8 & 95.4 &  69.4 & 83.9 & \underline{75.1} & \underline{63.3}     \\
            WaveMamba (ICCV'2025) & RGB+IR   & YOLOv8s  & 97.6 &  75.2 & 80.4 & 94.4 &  68.5 & 83.2 & -                & 62.9                \\
			\hline
			Ours (YOLOv8n)    & RGB+IR   & YOLOv8n   & \underline{98.3} &  76.2 & 80.7 & 96.5 &    69.5     &\underline{84.2}    &     74.7     & 62.1     \\
            Ours (YOLOv8s)    & RGB+IR   & YOLOv8s   & \textbf{98.5} &  \underline{77.0} & \textbf{82.3} & \textbf{97.5} & \textbf{71.8} & \textbf{85.4} & \textbf{75.7} & \textbf{63.4}     \\
            Ours (YOLOv13n)   & RGB+IR   & YOLOv13n  & \underline{98.3} &  69.2 & 73.5 & 96.1 &    67.5     &     81.0    &     71.3     & 59.3     \\
            Ours (YOLOv13s)   & RGB+IR   & YOLOv13s  & \underline{98.3} &  70.5 & 74.3 & 96.6 &    68.8     &     81.7    &     72.3     & 60.1     \\
			\hline
		\end{tabular}
    }
	\label{table1}
\end{table}

Table \ref{table2} evaluates the computational efficiency of different models. Several observations can be made. First, the lightweight variant, Ours-n (YOLOv8n), achieves the highest inference speed of 237.4 FPS on the validation set, indicating its potential for real-time inference. Second, M2I2HA provides a clear trade-off between accuracy and efficiency. Specifically, Ours-n (YOLOv8n) contains 6.1M parameters and achieves 84.2\% mAP@.5 and 62.1\% mAP on DroneVehicle, making it the efficiency-oriented variant. In contrast, Ours-s (YOLOv8s) achieves the best detection accuracy, with 85.4\% mAP@.5 and 63.4\% mAP, but requires 24.2M parameters and 68.8G FLOPs, making it the accuracy-oriented variant. Therefore, Ours-n is mainly used to demonstrate the efficiency advantage, whereas Ours-s represents the accuracy-oriented setting. Third, despite its lower theoretical GFLOPs, Ours-s (YOLOv13s) requires a higher training cost than Ours-s (YOLOv8s) (15.45 vs. 7.41 GPU-H). This may be attributed to the extensive use of depthwise separable convolutions (DSCs), which are often memory-bound on GPUs and may lead to fragmented memory access and reduced hardware utilization. 

\begin{table}[htbp]
    \centering
    \caption{Technical and efficiency indicators of different models on the DroneVehicle dataset.}
    \resizebox{0.8\textwidth}{!}{
        \begin{tabular}{@{}c|ccccc@{}}
            \hline
            Methods                 & Params (M)        & GFLOPs (G)        & FPS (Hz)            & GPU-H (h)          \\
            \hline
            CFT (arXiv'2022)        &     11.2          &     27.5          & 207.4               & \textbf{3.67}      \\
            SuperYOLO (TGRS'2023)   & \textbf{4.9}      & \underline{18.0}  & 122.9               & 7.24               \\
            GHOST (TGRS'2023)       &     7.1           &     20.5          & 165.8               & 10.42              \\
            ICAFusion (PR'2024)     &     23.3          &     30.1          & 101.1               & 14.50              \\
            GM-DETR (CVPR'2024)     &     69.8          &     172.8         & 80.7                & 13.25              \\
            COMO (IF'2025)          &     24.3          &     47.0          & \underline{226.5}   & \underline{5.25}   \\
            WaveMamba (ICCV'2025)   &     69.1          &     -             & 121.0               & 6.12               \\
            \hline
            Ours-n (YOLOv8n)        &     6.1           &     20.9          & \textbf{237.4}      & 5.45               \\
            Ours-s (YOLOv8s)        &     24.2          &     68.8          & 225.0               & 7.41               \\
            Ours-n (YOLOv13n)       &  \underline{5.2}  & \textbf{11.5}     & 212.5               & 7.26               \\
            Ours-s (YOLOv13s)       &     19.2          &     40.3          & 147.5               & 15.45              \\
            \hline
        \end{tabular}
    }
    \label{table2}
\end{table}

\begin{figure*}
    \centering
    \includegraphics[width=\textwidth]{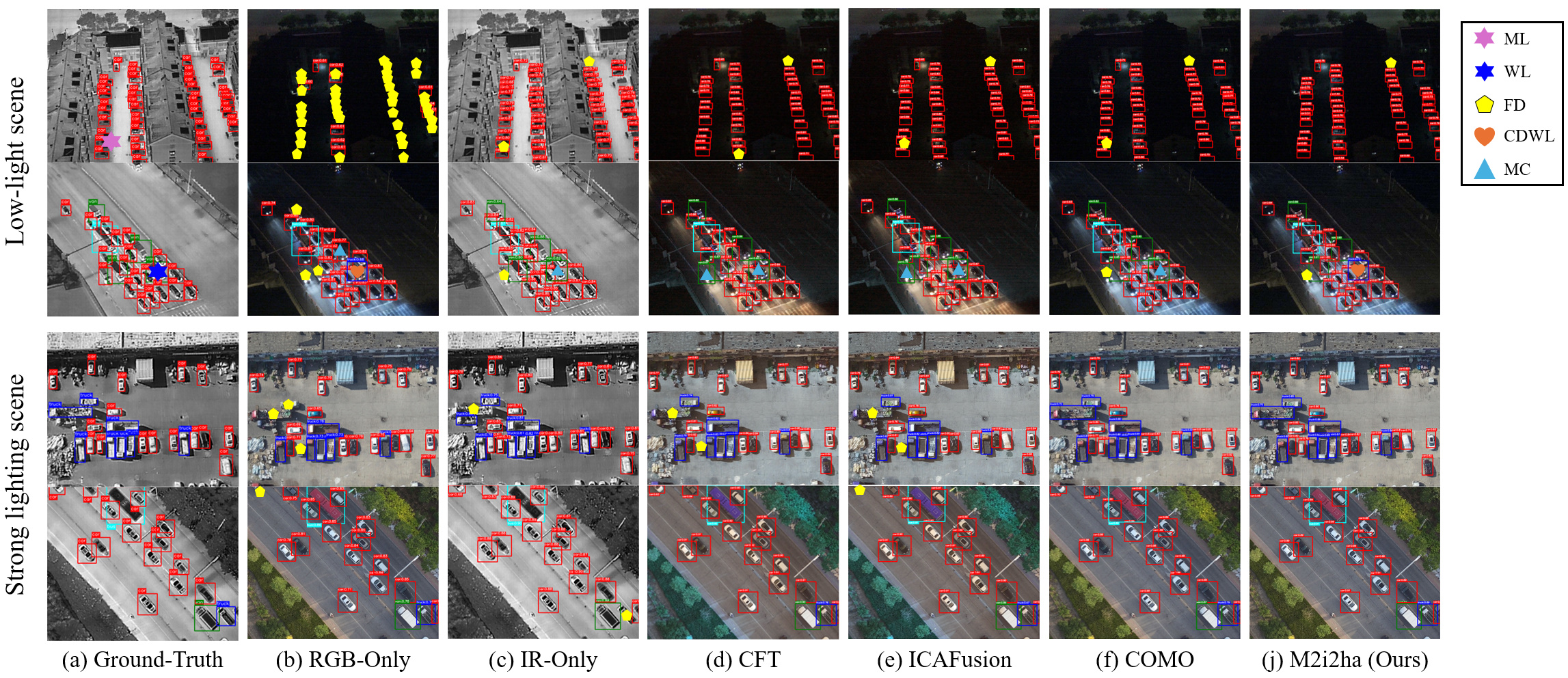}
    \caption{The detection results of different methods on the DroneVehicle dataset. In the figure, a purple hexagram denotes a Missing Label (ML) in the dataset, a blue hexagram represents a Wrong Label (WL) in the dataset, a yellow pentagon indicates a Failure Detection (FD), an orange heart shape marks a case of Correct Detection despite a Wrong Label (CDWL) in the dataset, and a blue triangle signifies a Misclassification (MC) in detection. It should be noted that objects which were not annotated in the dataset but were detected during the process are not marked particularly.}
    \label{fig7}
\end{figure*}

Fig. \ref{fig7} visualizes the multimodal detection results on RGB images against IR ground-truth. Single-modal methods suffer from severe missed detections, with RGB-only models failing under low light. Although multimodal methods improve performance, our approach achieves the lowest miss and false detection rates, and correctly identifies unlabeled or mislabeled objects, confirming the robustness of our feature enhancement and fusion strategy.

\subsection{Evaluation on the Flir-aligned Dataset}
The FLIR dataset (aligned version from ICAfusion \cite{shen2024icafusion}) is a dual-spectral benchmark featuring three annotated categories across 4129 training and 1013 testing RGB-IR pairs. Its limited scale strictly evaluates the data efficiency and robustness of models trained from scratch.

Table \ref{table3} compares performance after 60 training epochs. Single-modal infrared detection significantly outperforms visible-light detection, as frequent nighttime and complex lighting scenarios degrade RGB quality, whereas IR images offer superior environmental adaptability. Among multimodal methods, our YOLOv8s-based model achieves best results (72.5\% mAP@.5, 37.8\% mAP), underscoring its robust generalization on small-scale datasets. COMO remains highly competitive, securing the second-best results and slightly outperforming our method in mAP@.75. Unlike the dense, small-object scenarios in DroneVehicle, FLIR features driving perspectives with regular scene structures and larger object scales. In such context, COMO's scanning mechanism efficiently captures spatial context without complex relationship modeling. Conversely, our model maintains a distinct advantage on DroneVehicle, whose fine-grained feature fusion is optimized for fragmented and complex information. Wavemam performs equally well, with metrics comparable to those of Como.
\begin{table}[htbp]
	\centering
	\caption{Experiment results on the FLIR-aligned dataset.}
    \resizebox{0.85\textwidth}{!}{
		\begin{tabular}[htbp]{@{}c|ccc|ccc@{}}
			\hline
			Methods               & Person & Car     & Bicycle          & mAP@.5 (\%)      & mAP@.75 (\%)     & mAP (\%)                          \\
			\hline
			YOLOv8 (ADICS'2024)   & 57.0   &  79.5   & 36.9             &     58.0         &     20.2         & 26.4                              \\
			YOLOv8 (ADICS'2024)   & 77.1   &  87.7   & \underline{47.7} &     70.8         &     32.4         & 36.5                              \\
			YOLOv13 (arXiv'2025)  & 48.6   &  74.6   & 22.6             &     48.7         &     17.4         & 22.6                              \\
			YOLOv13 (arXiv'2025)  & 68.5   &  84.7   & 31.8             &     61.7         &     26.6         & 31.3                              \\
            \hline
			CFT (arXiv'2022)      & 73.5   &  86.2   & 44.9             &     68.2         &     27.5         & 32.6                              \\
            SuperYOLO (TGRS'2023) & 77.0   &  88.5   & 46.9             &     70.8         &     29.3         & 34.1                              \\
            GHOST (TGRS'2023)     & 75.6   &  84.5   & 46.6             &     68.9         &     28.4         & 30.6                              \\
            ICAFusion (PR'2024)   & 74.0   &  88.1   & 37.2             &     66.4         &     25.2         & 31.3                              \\
            GM-DETR (CVPR'2024)   & 76.2   &  88.3   & 46.1             &     70.2         &     29.1         & 34.6                              \\
            COMO (IF'2025)        & 77.8   &  87.9   & \textbf{49.4}    &     71.7         &  \textbf{34.0}   &\underline{37.5}                   \\
            WaveMamba (ICCV'2025) & 78.6   &  87.6   & 49.2             & \underline{71.8} &       -          & 37.4                              \\
			\hline
			Ours (YOLOv8n)        & \underline{78.9} & \textbf{89.3}    & 45.6             & 71.3            & 33.1             & 37.1            \\
            Ours (YOLOv8s)        & \textbf{81.0}    & \underline{88.9} & \underline{47.7} & \textbf{72.5}   & \underline{33.4} & \textbf{37.8}   \\
            Ours (YOLOv13n)       & 74.3             &  86.9            & 39.1             & 66.8            & 28.9             & 34.0            \\
            Ours (YOLOv13s)       & 74.2             &  86.6            & 41.3             & 67.4            & 30.2             & 34.8            \\
			\hline
		\end{tabular}
    }
	\label{table3}
\end{table}

\begin{figure*}
    \centering
    \includegraphics[width=\linewidth]{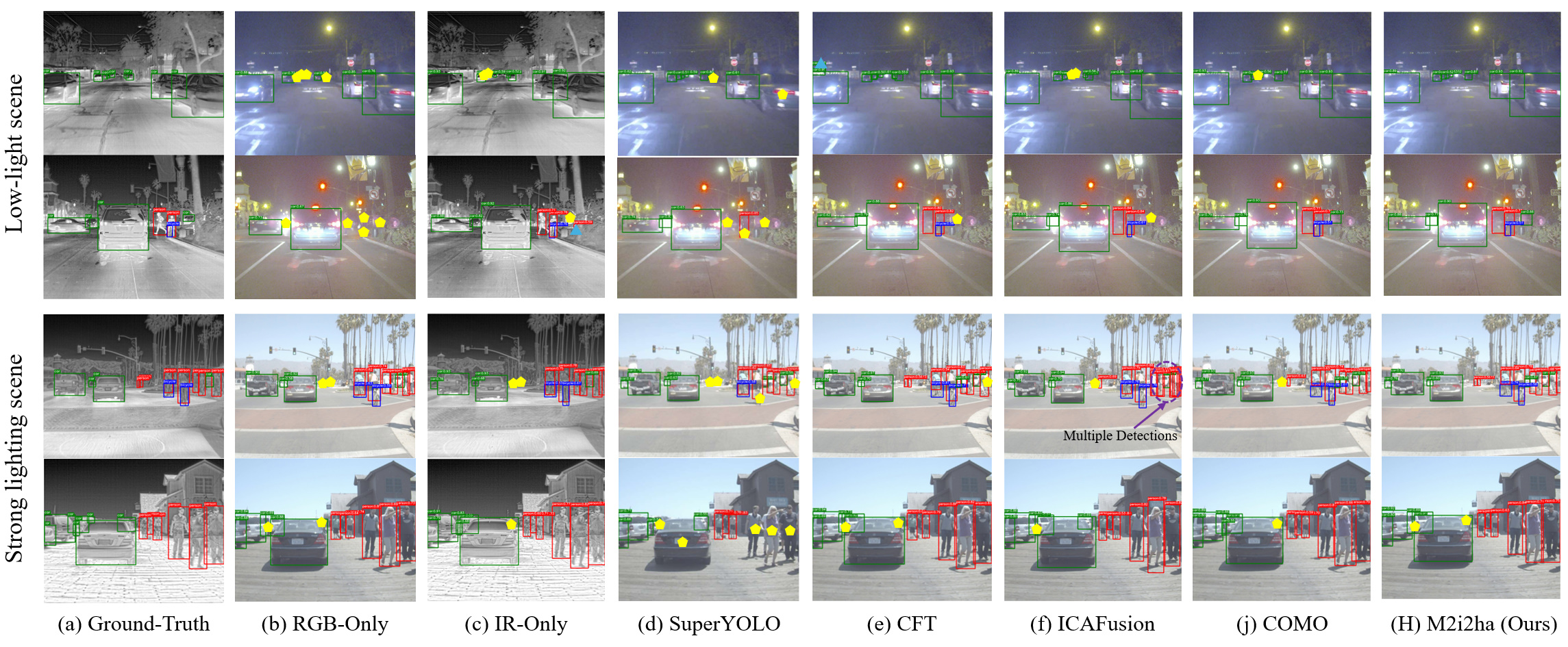}
    \caption{The detection results of different methods on the FLIR-aligned dataset.}
    \label{fig8}
\end{figure*}

Fig. \ref{fig8} visualizes detection results under challenging conditions (e.g., low illumination, glare, small targets, and partial occlusion). Notably, SuperYOLO successfully detects small objects but frequently misses larger ones. This indicates that its super-resolution task imposes excessive pixel-level constraints on large object regions, interfering with the extraction of semantic-level features essential for macro-structures.

\subsection{Evaluation on the LLVIP Dataset}
LLVIP is a visible-infrared benchmark for low-light vision tasks, comprising 15488 image pairs captured predominantly in extremely dark nocturnal environments. This presents a severe challenge to detection methods that rely heavily on visible-light features.

\begin{figure*}
    \centering
    \includegraphics[width=\textwidth]{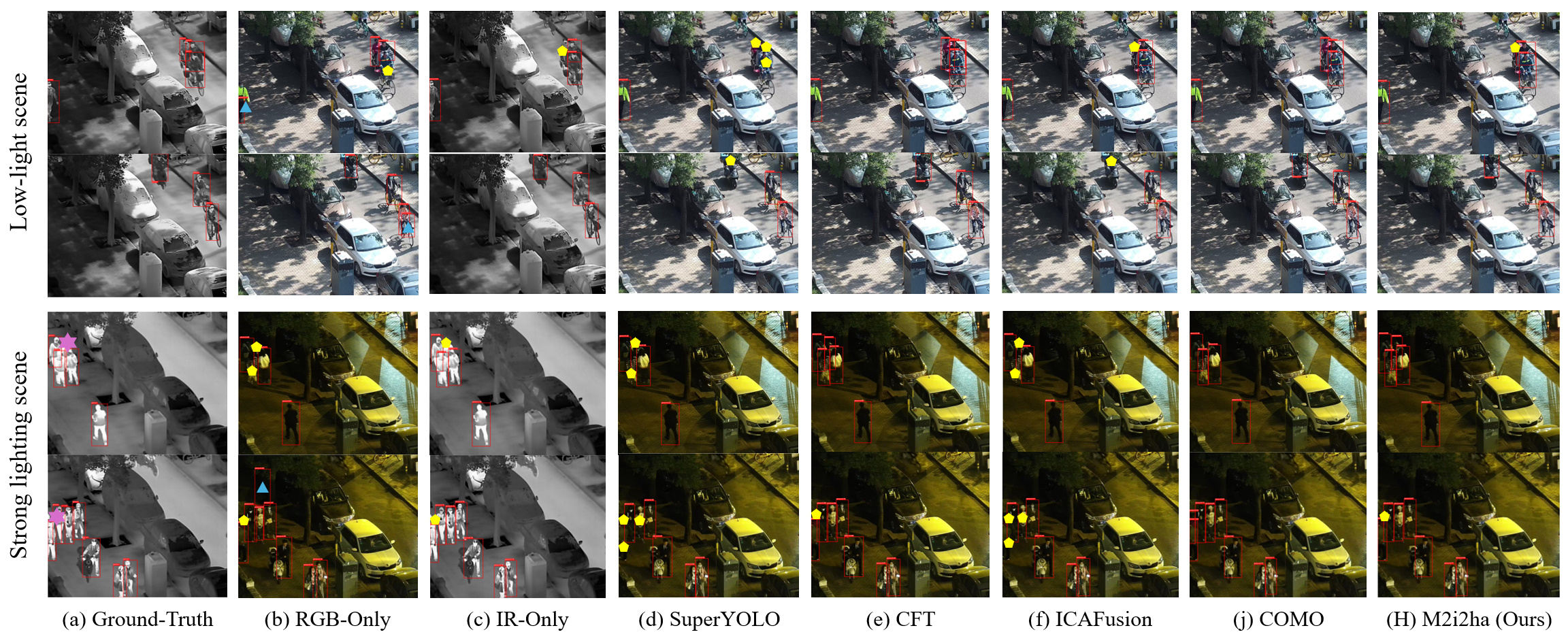}
    \caption{The detection results of different methods on the LLVIP dataset.}
    \label{fig9}
\end{figure*}

Table 4 presents the experimental results after 60 training epochs. Due to adverse lighting, the single-modal visible approach suffers a significant decline in detection accuracy. In contrast, our YOLOv8n-based multimodal model demonstrates an exceptional balance between accuracy and efficiency. It secures the second-best results in mAP@.5 (95.5\%) and overall mAP (59.7\%). Notably, on the stricter mAP@.75 metric, our model achieves best performance (68.0\%), confirming its superior localization precision and bounding box regression. Furthermore, our model requires only 1.9 GPU-hours for training, substantially outperforming most competitive methods. Among the baselines, CFT outperforms COMO and WaveMamba with the highest overall mAP and the lowest training cost (1.4 GPU-hours), indicating that its fusion strategy excels at exploiting infrared features under low-light conditions for rapid convergence. Qualitative comparisons of the detection results are visualized in Fig. \ref{fig9}.
\begin{table}[htbp]
	\centering
	\caption{Experiment results on the LLVIP dataset.}
    \resizebox{0.8\textwidth}{!}{
		\begin{tabular}[htbp]{@{}c|ccc@{}}
			\hline
			Methods                   & mAP@.5 (\%)  & mAP@.75 (\%)  & mAP (\%)     \\
			\hline
			YOLOv8 (ADICS'2024)       &     88.7               &     51.5                & 49.8                     \\
			YOLOv8 (ADICS'2024)       &     94.9               &     66.1                & 59.6                     \\
			YOLOv13 (arXiv'2025)      &     86.1               &     44.0                & 46.3                     \\
			YOLOv13 (arXiv'2025)      &     94.2               &     63.2                & 57.0                     \\
            \hline
			CFT (arXiv'2022)          &     95.2               &     \underline{66.5}    & \textbf{60.0}            \\
            SuperYOLO (TGRS'2023)     &     91.0               &     58.1                & 54.0                     \\
            GHOST (TGRS'2023)         &     93.7               &     62.5                & 58.9                     \\
            ICAFusion (PR'2024)       &     94.5               &     63.1                & 57.7                     \\
            GM-DETR (CVPR'2024)       &     94.7               &     60.2                & 55.7                     \\
            COMO (IF'2025)            &     \underline{95.5}   &     62.0                & 57.3                     \\
            WaveMamba (ICCV'2025)     &     95.3               &      -                  & 57.8                     \\
            \hline
			Ours (YOLOv8n)            &     \underline{95.5}   &     \textbf{68.0}       & \underline{59.7}         \\
            Ours (YOLOv8s)            &     \textbf{95.9}      &     65.6                & 58.8                     \\
            Ours (YOLOv13n)           &     94.7               &     59.2                & 55.4                     \\
            Ours (YOLOv13s)           &     95.3               &     65.5                & 58.9                     \\
			\hline
		\end{tabular}
    }
	\label{table4}
\end{table}

\subsection{Evaluation on the VEDAI Dataset}
\begin{figure*}
    \centering
    \includegraphics[width=\textwidth]{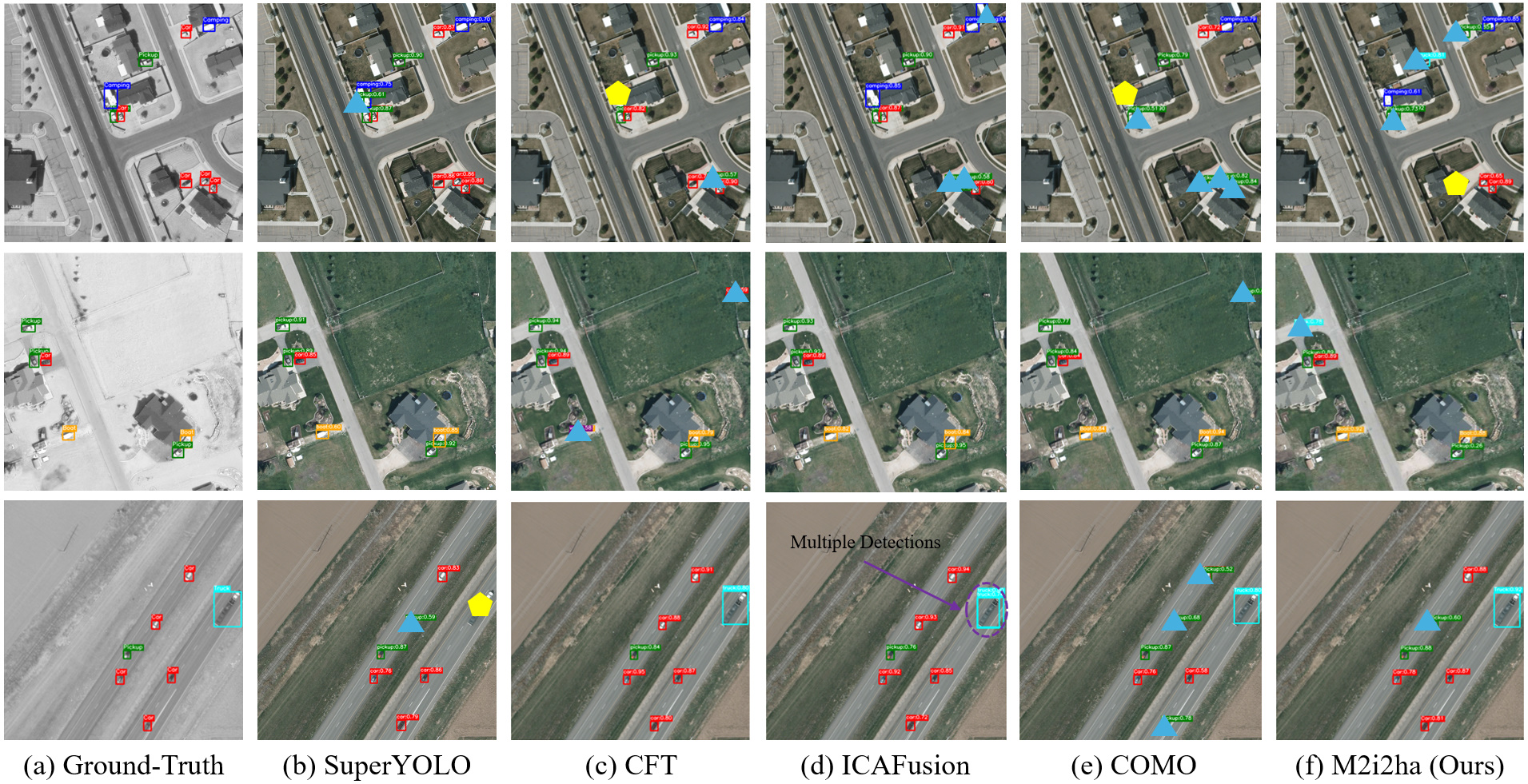}
    \caption{The detection results of different methods on the VEDAI dataset.}
    \label{fig10}
\end{figure*}

VEDAI is an aerial detection benchmark with minute objects and complex backgrounds. We adopt the split from SuperYOLO \cite{zhang2023superyolo} (1089 training, 118 testing pairs) with data augmentation and a 300-epoch training scheme. Input resolution is $1024 \times 1024$ for SuperYOLO to support its super-resolution task, and $512 \times 512$ for all other methods.As shown in Table 5, SuperYOLO achieves SOTA performance (75.4\% mAP@.5, 47.5\% mAP) due to its multimodal Super-Resolution (SR) auxiliary branch, which reconstructs high-frequency details to resolve few-pixel targets. COMO ranks second in overall mAP (44.2\%), closely followed by our model at 43.6\% mAP, which also secures the second-best mAP@.5 (74.5\%). This competitiveness in data-scarce remote sensing scenarios is driven by our Hypergraph-based fusion and M3FDFP block, which adaptively preserve critical low-level structural features during multimodal integration. This ensures robust detection capabilities even for minute targets and limited training data. Figure \ref{fig10} illustrates the comparative detection results.

\begin{table}[htbp]
	\centering
	\caption{Experiment results on the VEDAI dataset.}
    \resizebox{0.7\textwidth}{!}{
		\begin{tabular}[htbp]{@{}c|cccccccc|ccc@{}}
			\hline
			Methods               & mAP@.5 (\%)  & mAP (\%)  & GPU-H (h)  \\
			\hline
			YOLOv8 (ADICS'2024)   &     63.3    & 37.5      & -        \\
			YOLOv8 (ADICS'2024)   &     59.4    & 36.0      & -        \\
            \hline
			CFT (arXiv'2022)      &     63.6    & 35.4      & \textbf{0.52}     \\
            SuperYOLO (TGRS'2023) &\textbf{75.4}    & \textbf{47.5}      & 5.10     \\
            ICAFusion (PR'2024)   &     63.9    & 38.2      & 3.46     \\
            COMO (IF'2025)        &     71.0    & \underline{44.2}      & \underline{0.93}     \\
			\hline
            Ours (YOLOv8s)        & \underline{74.5}    & 43.6     & 1.31     \\
			\hline
		\end{tabular}
    }
	\label{table5}
\end{table}

\subsection{Ablation Studies}
\textbf{Overall Module Ablations.} To systematically evaluate the core components of the proposed M2I2HA framework, we conduct progressive ablation studies on the DroneVehicle dataset using a YOLOv8s backbone. The evaluation focuses on three key components: the Intra-Hypergraph Enhancement (IHE) module, the Inter-Hypergraph Fusion (IHF) module, and the M3FDFP block. A simple convolutional fusion structure following COMO \cite{liu2025cross} serves as the multi-modal baseline.

\begin{figure}
    \centering
    \includegraphics[width=0.5\textwidth]{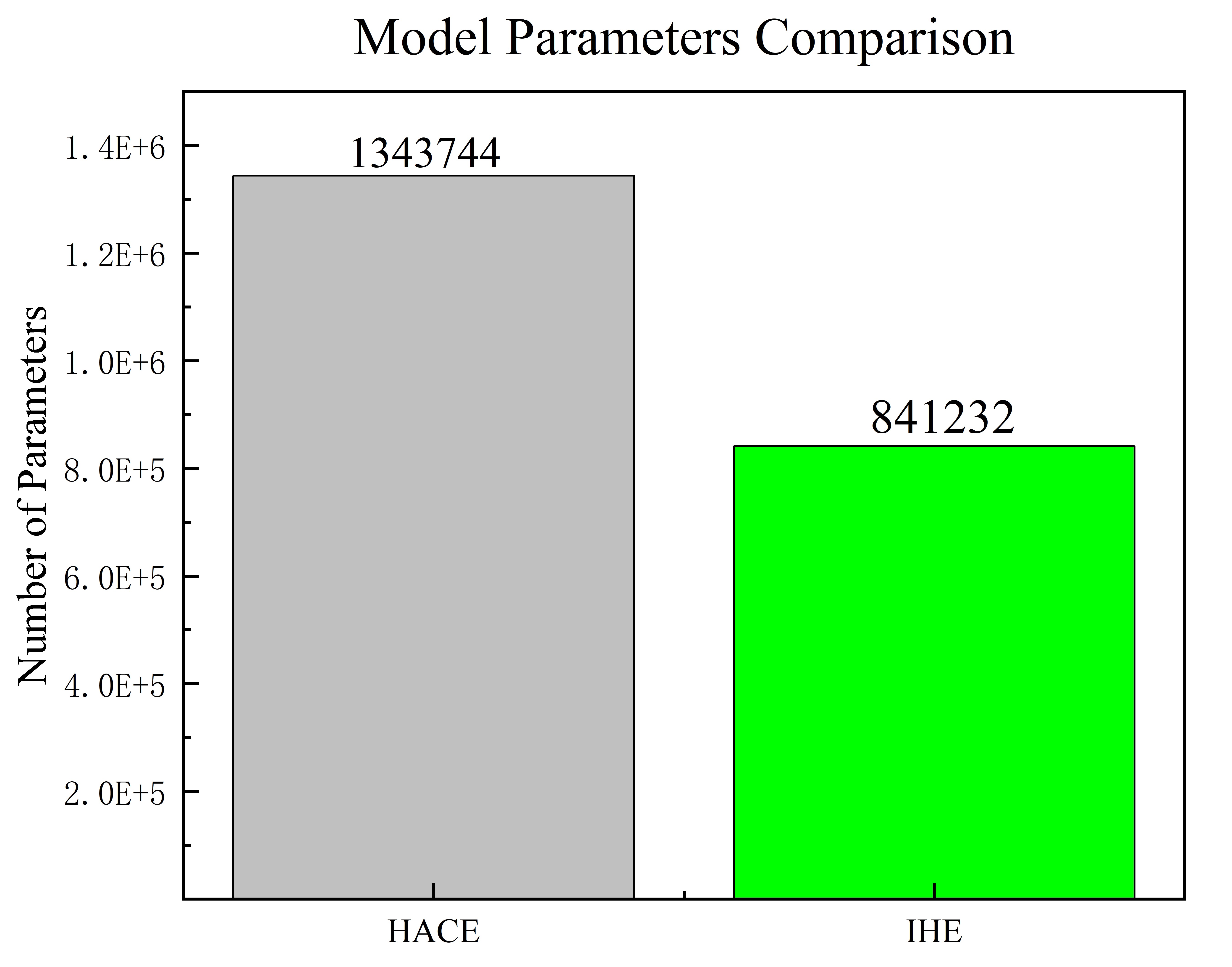}
    \caption{Parameter Count: HACE vs. IHE.}
    \label{fig11}
\end{figure}

As detailed in Table \ref{table6}, the multi-modal baseline naturally outperforms single-modal counterparts due to complementary information. Integrating the original HyperACE module (``Baseline+HACE'') improves mAP@.5 and mAP by 1.4\% and 1.2\%, respectively. Replacing it with our improved IHE module further enhances detection accuracy while simultaneously reducing the module parameter count by 37.4\% (Figure \ref{fig11}). Notably, ``Baseline+IHF'' surpasses ``Baseline+IHE'', demonstrating that inter-modal fusion exerts a more critical impact on performance than intra-modal enhancement. Finally, incorporating the M3FDFP block brings a substantial performance leap, owing to its dynamic mechanism that effectively preserves low-level features for small object detection. The complete framework yields the optimal results, fully validating the synergy of the proposed modules.
\begin{table}[htbp]
    \centering
    \caption{Ablation Studies of different modules on the DroneVehicle dataset.}
    \resizebox{0.6\textwidth}{!}{%
        \begin{tabular}{@{}c|cccccc@{}}
            \hline
            Methods        & HACE   & IHE   & IHF   & M3FDFP  & mAP.5 (\%)  & mAP (\%)     \\
            \hline
            RGB-only                                     &        &       &       &       &        72.8           &          44.8          \\
            IR-only                                      &        &       &       &       &        78.5           &          57.1          \\
            Baseline                                     &        &       &       &       &        80.2           &          58.3          \\
            $\text{Baseline} + \text{HACE}$              &$\surd$ &       &       &       &        81.6           &          59.5          \\
            $\text{Baseline} + \text{IHE}$               &        &$\surd$&       &       &        81.9           &          60.8          \\
            $\text{Baseline} + \text{IHF}$               &        &       &$\surd$&       &        82.2           &          61.9          \\
            $\text{Baseline} + \text{IHE} + \text{M3FDFP}$ &        &$\surd$&       &$\surd$&        82.0           &          61.2          \\
            $\text{Baseline} + \text{IHF} + \text{M3FDFP}$ &        &       &$\surd$&$\surd$&        82.7           &          62.3          \\
            $\text{Baseline} + \text{IHE} + \text{IHF}$  &        &$\surd$&$\surd$&       &    \underline{83.9}   &     \underline{62.6}   \\
            \hline
            Ours                                         &        &$\surd$&$\surd$&$\surd$&        \textbf{85.4}  &          \textbf{63.4} \\
            \hline
        \end{tabular}%
    }
    \label{table6}
\end{table}

\textbf{IHF Ablations.}
To evaluate the effectiveness of the proposed IHF module for multimodal fusion, we conduct ablation experiments on the DroneVehicle dataset using a YOLOv8s backbone within the existing M2I2HA framework, where IHF is replaced with alternative fusion strategies. Specifically, we consider six configurations: (1) Replacing IHF with a fusion module based on successive addition and convolution operations. (2) Replacing IHF with a plain convolutional fusion module based on simple concatenation and convolution. (3) Replacing IHF with a spatial-channel gated fusion module \cite{tao2023spatial}. (4) Replacing IHF with a Transformer-based cross-attention (CrossAtten) mechanism \cite{chandrasiri2023cross}. (5) Replacing IHF with a MambaFusion module in COMO \cite{liu2025cross}. (6) Optimizing the original IHF by incorporating low-rank decomposition of the node matrix.  The results are reported in Table \ref{table7}. As shown, the M2I2HA(IHF) consistently outperforms the other variants in terms of both mAP@0.5 and mAP. This demonstrates that the IHF module possesses a stronger ability to capture cross-modal feature relationships, thereby effectively boosting the performance of multi-modal object detection. 
\begin{table}[htbp]
    \centering
    \caption{Ablation Studies of different cross-modal fusion modules on the DroneVehicle dataset.}
    \resizebox{0.6\textwidth}{!}{
        \begin{tabular}{@{}c|cccc@{}}
            \hline
            Methods                       &  Params (M)    &  GFLOPs (G)      & mAP.5 (\%)    & mAP (\%)      \\
            \hline
            M2I2HA(Add+Conv)              &  20.1          &  67.1            & 81.4          & 60.1          \\
            M2I2HA(Concate+Conv)          &  20.3          &  67.2            & 81.9          & 60.8          \\
            M2I2HA(Spatial-Channel Gated) &  20.7          &  67.8            & 82.5          & 61.7          \\
            M2I2HA(CrossAtten)            &  21.5          &  67.8            & 83.9          & 62.3          \\
            M2I2HA(MambaFusion)           &  21.6          &  67.4            & 84.2          & 62.5          \\
            M2I2HA(IHF+Node Low-Rank)     &  20.8          &  67.6            & 83.9          & 61.8          \\
            M2I2HA(IHF)                   &  24.2          &  68.8            & \textbf{85.4} & \textbf{63.4} \\
            \hline
        \end{tabular}
    }    
    \label{table7}
\end{table}

\subsection{Analysis of the Dynamic Routing of M3FDFP Block}
\begin{figure}
    \centering
    \includegraphics[width=0.9\textwidth]{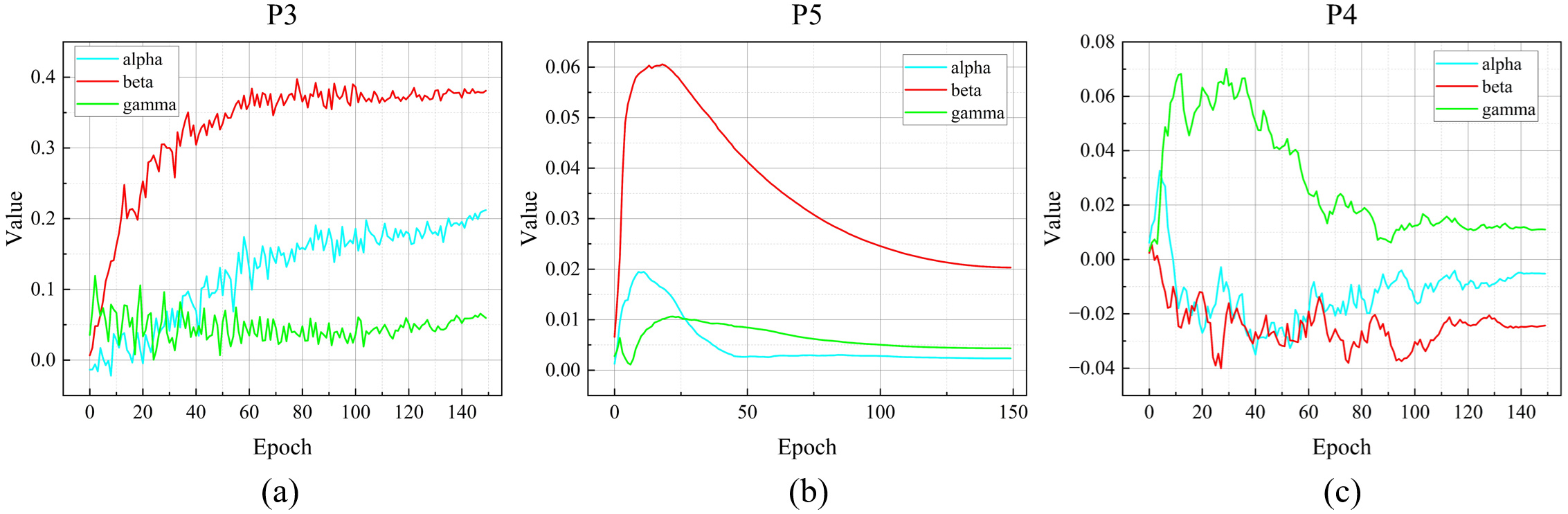}
    \caption{Dynamic changes of the learnable routing parameters in the M3FDFP Block. It should be noted that these coefficients are learnable residual scalars rather than normalized fusion probabilities.}
    \label{fig12}
\end{figure}

Fig. \ref{fig12} illustrates the training evolution curves of the learnable scalars ($\alpha$, $\beta$, and $\gamma$) within the P3-P5 M3FDFP blocks. Several key insights can be observed: (1) The weights of $\alpha$, $\beta$, and $\gamma$ exhibit distinct distributions across different feature scales; (2) The M3FDFP block successfully implements dynamic routing for various augmented features throughout training; (3) All feature branches maintain non-trivial contributions upon convergence without degenerating into a single dominant source, thereby validating the architectural necessity and effectiveness of each individual module.

\subsection{Analysis of the Hyperedge Construction}
\begin{figure}
    \centering
    \includegraphics[width=0.9\textwidth]{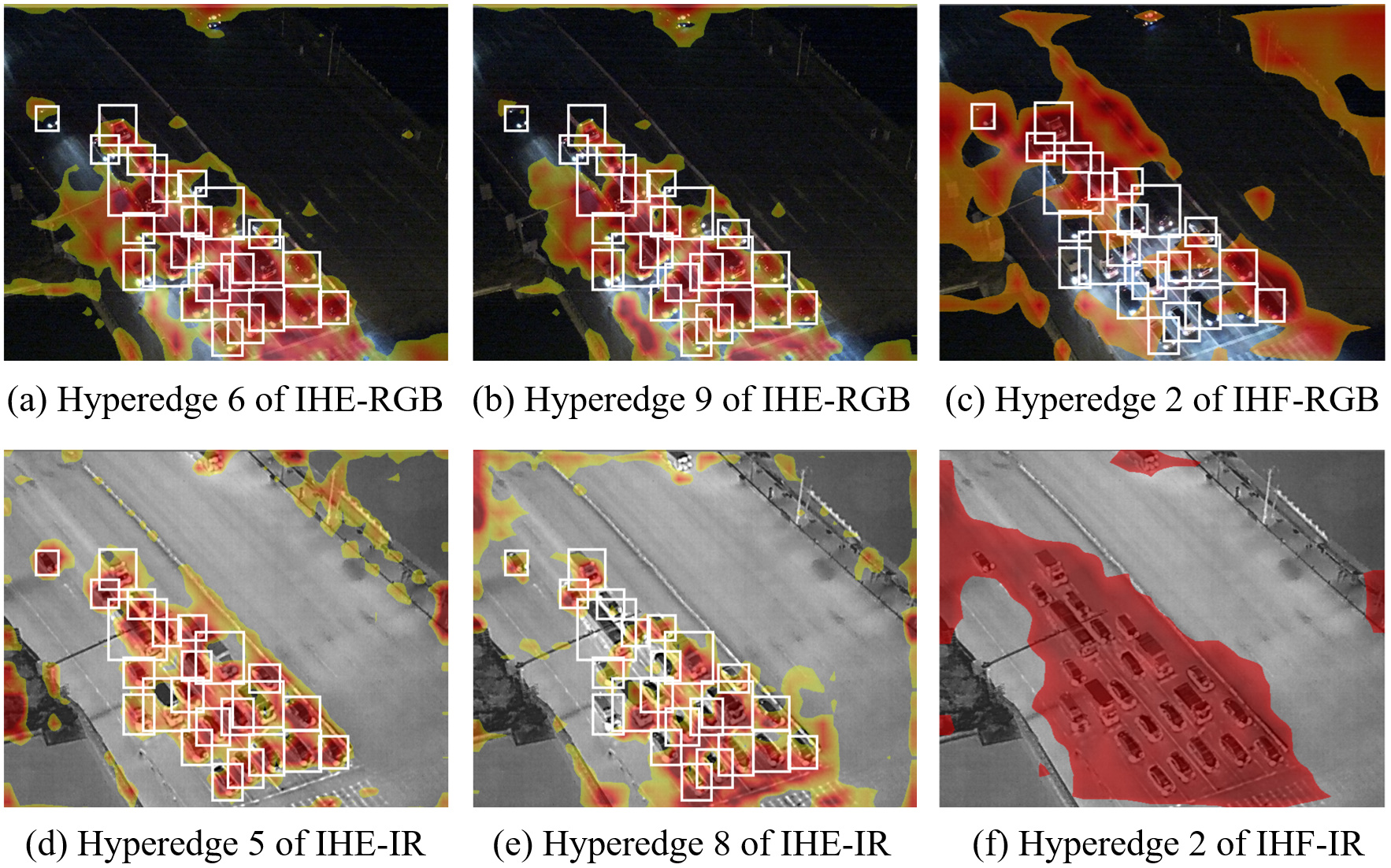}
    \caption{Visualization of hyperedge construction in the M2I2HA network. The heatmaps are obtained by projecting the node-hyperedge assignment weights of representative learned hyperedges back to the input image space. Representative hyperedges are selected according to their object-region response within ground-truth boxes.}
    \label{fig13}
\end{figure}

Fig. \ref{fig13} visualizes representative hyperedge responses from the trained IHE and IHF modules via heatmap overlays on RGB and IR images, yielding two key observations. First, the intra-modal hypergraphs capture meaningful non-local, group-wise relationships rather than merely local neighborhoods; specifically, they inherently group spatially separated object regions and their surrounding contextual cues within each modality. Second, the inter-modal hypergraph successfully facilitates complementary cross-modal exchange. In low-illumination scenarios, the hyperedge selectively attends to weak visual cues (e.g., vehicle bodies and headlights) on the RGB side, while simultaneously covering distinct thermal object regions on the IR side. This demonstrates that IHF effectively establishes cross-modal group associations to enhance modality-complementary representation learning. 

\subsection{Analysis of the Attention Mechanism}
\begin{figure}
    \centering
    \includegraphics[width=0.9\textwidth]{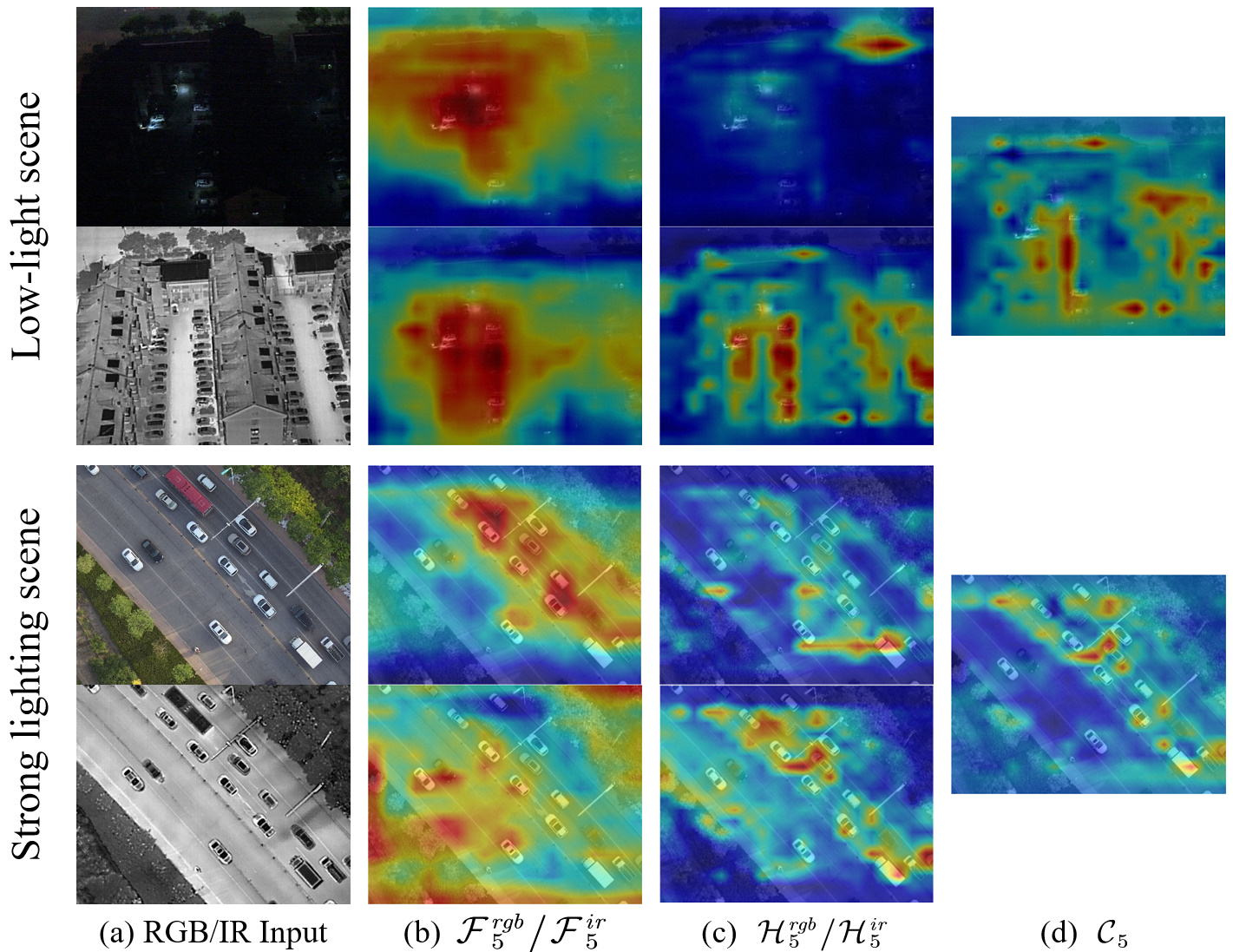}
    \caption{Visualization of attention maps at different stages.}
    \label{fig14}
\end{figure}

Fig. \ref{fig14} visualizes the multi-stage attention maps of the M2I2HA framework across varying illumination conditions. Initial backbone features (b) are relatively diffused, with the RGB branch exhibiting pronounced background noise in low-light scenes. Progressively, the IHE module (c) suppresses this noise and reinforces discriminative intra-modal semantic features. Under inter-hypergraph fusion, the cross-modal information interplay allows salient IR thermal features to effectively guide and ``repair'' degraded RGB representations, yielding highly aligned cross-modal spatial distributions. Consequently, the final fused attention maps (d) display precise target localization and sharp boundary contours. Furthermore, cross-lighting comparisons reveal that the framework dynamically shifts its reliance toward RGB inputs in well-illuminated environments. This adaptive attention allocation demonstrates the network's capacity to adjust modality weights based on environmental characteristics rather than biasing toward a single dominant modality, successfully mitigating information redundancy and overreliance.

\section{Conclusions and Future Works}
\label{sec5}
This paper proposes M2I2HA, a novel hypergraph attention-based multi-modal object detection framework designed for robust target recognition under challenging conditions. M2I2HA incorporates two core components: the Intra-Hypergraph Enhancement (IHE) and Inter-Hypergraph Fusion (IHF) modules, which model and strengthen fine-grained feature relationships within and across modalities, respectively. Furthermore, an adaptive dynamic feature allocation and fusion mechanism is introduced to mitigate deep-network information bottlenecks and enhance resilience against environmental noise. Extensive experiments on public datasets demonstrate that M2I2HA is highly competitive against the current state-of-the-art methods. Crucially, it achieves superior accuracy with lower computational overhead, making it highly suitable for real-time edge applications. Future work will extend M2I2HA to broader modalities and advanced lightweight backbones.

\bibliographystyle{elsarticle-num}
\bibliography{reference}

\end{document}